\documentclass[sigconf]{acmart}
\copyrightyear{2021}
\acmYear{2021}
\setcopyright{acmlicensed}\acmConference[WSDM '21]{Proceedings of the Fourteenth ACM International Conference on Web Search and Data Mining}{March 8--12, 2021}{Virtual Event, Israel}
\acmBooktitle{Proceedings of the Fourteenth ACM International Conference on Web Search and Data Mining (WSDM '21), March 8--12, 2021, Virtual Event, Israel}
\acmPrice{15.00}
\acmDOI{10.1145/3437963.3441810}
\acmISBN{978-1-4503-8297-7/21/03}

\usepackage[utf8]{inputenc} 
\usepackage[T1]{fontenc}    
\usepackage{hyperref}       
\usepackage{url}            
\usepackage{booktabs}       
\usepackage{amsfonts}       
\usepackage{nicefrac}       
\usepackage{microtype}      
\usepackage{algorithm}
\usepackage{xspace}
\usepackage{multirow}
\usepackage{dsfont}
\usepackage{amsmath} 
\usepackage{graphicx}
\usepackage{mathtools}
\usepackage{purudefs}
\usepackage{subcaption}
\usepackage{longtable}

\newcommand{\suppl}{\href{http://manikvarma.org/pubs/dahiya21.pdf}{\color{blue}{supplementary material}}\xspace}
\newcommand{\code}{\href{https://github.com/Extreme-classification/deepxml}{\color{blue}{https://github.com/Extreme-classification/deepxml}}\xspace}
\newcommand{\fm}{DeepXML\xspace}
\newcommand{\alg}{Astec\xspace}

\newtheorem{theorem}{Theorem}

\begin{document}
\fancyhead{}

\title{DeepXML: A Deep Extreme Multi-Label Learning Framework Applied to Short Text Documents}

\author{Kunal Dahiya}
\email{kunalsdahiya@gmail.com}
\affiliation{%
  \institution{IIT Delhi}
  \country{India}
}

\author{Deepak Saini}
\email{desaini@microsoft.com}
\affiliation{%
  \institution{Microsoft Research}
  \country{India}
}

\author{Anshul Mittal}
\email{me@anshulmittal.org}
\author{Ankush Shaw}
\email{shawank17198@gmail.com}
\affiliation{%
  \institution{IIT Delhi}
  \country{India}
}

\author{Kushal Dave}
\email{kudave@microsoft.com}
\author{Akshay Soni}
\email{Akshay.Soni@microsoft.com}
\affiliation{%
  \institution{Microsoft}
  \country{USA}
}

\author{Himanshu Jain}
\email{himanshu.j689@gmail.com}
\author{Sumeet Agarwal}
\orcid{0000-0002-5714-3921}
\email{sumeet@ee.iitd.ac.in}
\affiliation{%
  \institution{IIT Delhi}
  \country{India}
}

\author{Manik Varma}
\email{manik@microsoft.com}
\affiliation{%
    \institution{Microsoft Research}
    \institution{IIT Delhi}
    \country{India}
}

\renewcommand{\shortauthors}{K. Dahiya et al.}



\begin{abstract}
Scalability and accuracy are well recognized challenges in deep extreme multi-label learning where the objective is to train architectures for automatically annotating a data point with the most relevant subset of labels from an extremely large label set. This paper develops the \fm framework that addresses these challenges by decomposing the deep extreme multi-label task into four simpler sub-tasks each of which can be trained accurately and efficiently. Choosing different components for the four sub-tasks allows \fm to generate a family of algorithms with varying trade-offs between accuracy and scalability. In particular, \fm yields the \alg algorithm that could be 2-12\% more accurate and 5-30$\times$ faster to train than leading deep extreme classifiers on publically available short text datasets. \alg could also efficiently train on Bing short text datasets containing up to 62 million labels while making predictions for billions of users and data points per day on commodity hardware. This allowed \alg to be deployed on the Bing search engine for a number of short text applications ranging from matching user queries to advertiser bid phrases to showing personalized ads where it yielded significant gains in click-through-rates, coverage, revenue and other online metrics over state-of-the-art techniques currently in production. DeepXML's code is available at \code.
\end{abstract}
\maketitle
\section{Introduction}

\textbf{Objective}: This paper develops the \fm framework for deep extreme multi-label learning where the goal is to train architectures for automatically annotating a data point with the most relevant {\it subset} of labels from an extremely large label set. Note that multi-label learning generalizes multi-class classification which aims to predict a single mutually exclusive label. The objectives for developing \fm are threefold. First, \fm provides a framework for how to think about deep extreme multi-label learning that can not only be used to analyze seemingly disparate algorithms such as XML-CNN~\cite{Liu17} and MACH~\cite{Medini19} but which can also be used to derive significantly improved versions of such state-of-the-art deep extreme classifiers. Second, \fm can generate a family of new state-of-the-art algorithms obtained by combining various types of feature architectures with different classifiers in a scalable and accurate manner. In particular, \fm was used to derive \alg (standing for an Accelerated Short Text Extreme Classifier) which is specialized for extremely low-latency and high throughput short text applications as it can make billions of predictions per day and handle peak rates of up to a hundred and twenty thousand queries per second. Third, \fm provides an easy-to-use modular framework in which practitioners can design architectures for diverse applications by making minimal changes and simply plugging in the components of their choice rather than going back to the drawing board and designing a specialized architecture for each application from scratch.

\textbf{Short text applications}: This paper focuses on the classification of short text documents, having just 3-10 words on average, into millions of labels. In particular, this paper considers a range of short text applications including matching user search engine queries to advertiser bid phrases, predicting Wikipedia tags from a document's title, predicting frequently bought together products from a given retail product's name and showing personalized ads based on the set of webpage titles in a user's browsing history. Such applications pose a number of additional challenges to deep extreme classifiers as compared to long text documents having up to 200 words on average. First, the deep extreme classifier is forced to make predictions on the basis of just 3-10 words on average which is a significantly harder task than long text document classification. Second, short text corpora have fewer occurrences of each word than long text corpora thereby leading to a paucity of training data. For instance, the Bing datasets have millions of words which occur at most twice in the training set and thus learning good quality embeddings for such rare words can be challenging. Third, low-latency and high-throughput short text applications with billions of users require predictions in milliseconds on a CPU to keep operating and energy costs low. While these challenges seem daunting at the extreme scale, it is nevertheless important to design solutions for short text applications as they can benefit billions of people.

\textbf{Challenges in deep extreme classification}: Deep extreme classifiers jointly learn a feature architecture with an extremely large  classification layer leading to the following challenges. First, training and fine-tuning the feature architecture for millions of labels can be computationally expensive and can also lead to learning poor quality representations when training data is scarce. Second, both the forward prediction pass as well as gradient back-propagation become infeasible if the classification layer has costs that are linear in the number of labels (such as for a fully connected output layer). One might be tempted to address these challenges by replacing the classification layer with highly scalable non-deep learning based extreme classifiers~\cite{Prabhu18b, Jain19} which reduce the costs of the forward and backward pass to logarithmic in the number of labels. This is achieved by learning a sub-linear search data structure based on graphs~\cite{Jain19}, trees~\cite{Prabhu18b,Khandagale19,Wydmuch18,Prabhu14,Jain16,Prabhu18}, hashes~\cite{Tagami17,Siblini18a} or clustering~\cite{Bhatia15} in a fixed feature space such as a bag-of-words representation or fixed pre-trained embeddings. Unfortunately, such algorithms cannot be directly used to replace the classification layer in the deep learning setting as the feature representation changes with every mini-batch update. This necessitates the frequent recomputation or updation of the sub-linear search data structure on the updated features which can be prohibitively expensive. As such, deep learning presents additional statistical and computational challenges to extreme classification.

\textbf{\fm}: \fm addresses these challenges by decomposing the deep extreme learning task into the following four sub-tasks or modules each of which can be constrained to be learnt in log-time while maintaining accuracy. In Module I, an intermediate feature representation is learnt using an application-appropriate feature architecture trained on a simpler surrogate task. The objective in Module I is to efficiently learn a near-final feature representation which can be fixed and used to eliminate all but a logarithmic number of the hardest negative labels for each data point. Then, in Module II, a graph, tree, hash or clustering based sub-linear search data structure is trained just once on the fixed intermediate features to shortlist the hardest negative labels for each data point in log-time. The motivation is to reduce the problem for any given data point from an extreme task with millions of labels to a traditional classification task with just hundreds of labels. Note that this can be achieved with minimal loss in accuracy as uninformative negative labels can be discarded since they don't contribute to the final solution~\cite{Jain19}. Module III then transfers the intermediate features to learn final features for the given extreme task subject to the constraint that the final features don't lie very far away from the intermediate features. This is done so as to get all the accuracy gains of fine-tuning for the task at hand while ensuring that the hardest negatives continue to lie in the shortlisted label set. Module IV jointly learns an extreme classifier along with the final features in log-time using just the shortlisted labels. Varying the choices for the component including the feature architecture, the surrogate task, the sub-linear search data structure, the negative sampling procedure, the transfer mechanism and the extreme classifier leads to a family of state-of-the-art algorithms including \alg, DECAF~\cite{Mittal21}, GalaXC~\cite{Saini21}, ECLARE~\cite{Mittal21b}, {\it etc}.

\textbf{\alg}: \alg was derived from \fm specifically for short text applications. It employed a low capacity feature architecture which could be learnt accurately from limited training data. This allowed \alg to be 2-12\% more accurate than leading deep extreme classifiers on publicly available benchmark datasets while also being up to 20\% more accurate than state-of-the-art techniques for matching user queries to advertiser bid phrases on Bing datasets. Furthermore, by leveraging the \fm framework, \alg could be 5-30$\times$ faster to train than leading deep extreme classifiers and could efficiently scale to problems involving 62 million labels. Finally, \alg could make predictions in milliseconds on a CPU and could therefore make billions of predictions per day, with peak rates of 120,000 queries per second, using commodity hardware. As a result, \alg increased the click-through-rate by 6.5\% over state-of-the-art techniques in production for showing personalized ads based on the webpage titles and URLs in a user's browsing history. Similarly, \alg yielded an increase of 1.6\% in revenue per thousand queries, a 2.9\% increase in match quality and an 8.6\% increase in query coverage over leading techniques in production for matching user queries to advertiser bid phrases. These represent just two of the multiple short text applications for which \alg resulted in a significant increase in key metrics on Bing.

\section{Related Work} \label{sec:lit}

\textbf{\fm and extreme Classification}: Much progress has been made in extreme classification~\citep{Agrawal13,Prabhu14,Mineiro15,Jain16,Jain19,Yen16,Yen18a,Babbar17,Babbar19,Prabhu18,Prabhu18b,Khandagale19,Bhatia15,Tagami17,Wydmuch18,Jasinska16} for fixed representations such as bag-of-words features and pre-trained embeddings. Unfortunately, these algorithms cannot be used directly for deep extreme classification as the sub-linear search data structure they rely on for scalability needs to be frequently recomputed due to the change in features with every mini-batch update. As a result, specialized deep extreme classifiers~\citep{Liu17, Medini19, You18, Jernite17, Ye20, Yuan20, Chalkidis19} have been developed and many of these can be analysed and improved in the \fm framework.

\textbf{\alg and short text extreme classification}: Of particular relevance to \alg and this paper are specialized extreme classifiers that have been developed for short text applications including Slice~\citep{Jain19} for recommending related queries on Bing and MACH~\citep{Medini19} for searching Amazon retail products. Slice is a highly scalable classifier for fixed pre-trained embeddings and, unfortunately, cannot be used for deep extreme classification as already mentioned. \alg could therefore be up to 20\% more accurate than Slice on pre-trained CDSSM~\citep{Huang13}, FastText~\citep{Joulin17} or BERT~\citep{Devlin19} embeddings (see Section~\ref{sec:results}). Also note that Slice could always be incorporated into \fm's second module if desired. Furthermore, \alg can be seen as a generalization of MACH when analyzed through the \fm framework. In particular, MACH stops training after the first \fm module and therefore has to learn a large ensemble of base classifiers to compensate. \alg could be up to 12\% more accurate and orders of magnitude faster to train as it learnt a single base classifier by efficiently and accurately leveraging the \fm  framework with all four modules.

\begin{figure}
    \centering
    \includegraphics[width=0.92\columnwidth, clip]{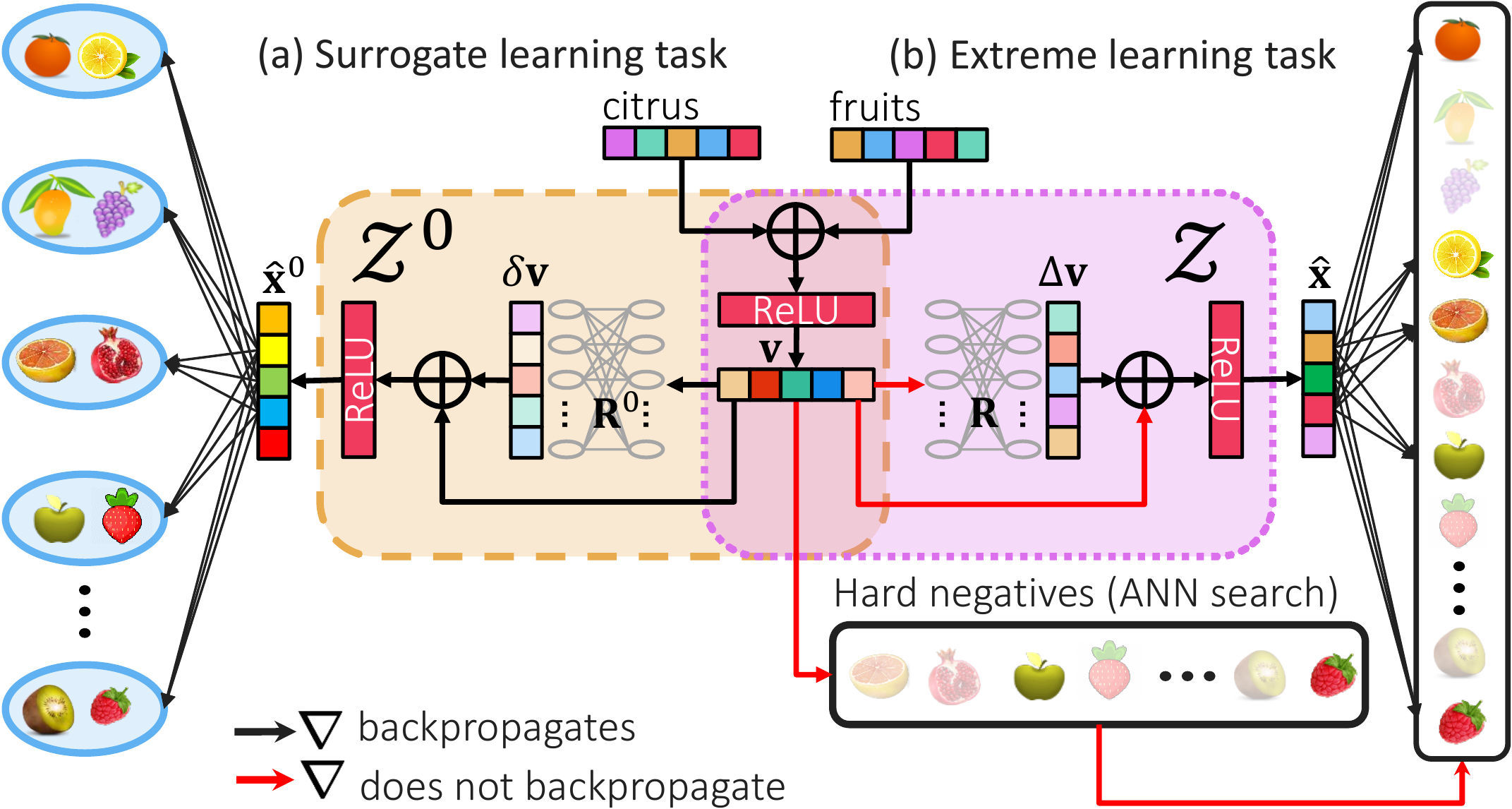}
    \caption{\alg's architecture: Please refer to the text for details. Best viewed under magnification and in color.}
    \label{fig:arch}
\end{figure}

\textbf{Short text applications}: Apart from extreme classification, four classes of techniques have been developed for matching user queries to advertiser bid phrases and the other short text applications considered in this paper. The first class of techniques leverage additional sources of information such as landing pages~\cite{Yejin10}, web search results and other queries~\cite{Broder09} and are therefore beyond the scope of this paper. The second class of techniques are based on generative models~\citep{Gao12,Zhou19,Lee18,Lian19} that synthesize bid phrases for a given query. Unfortunately, unconstrained synthesis can result in poor quality bid phrases being generated while constraining them through tries~\cite{Lian19} or other approaches can limit bid phrase coverage. The third class of techniques use graph neural networks, random walks and other graph processing methods on the query-bid phrase, query-url, query-token or query-query graphs~\cite{Ioannis08,Mei08}. Features based on such graphs can be readily incorporated into the \fm framework by leveraging graph neural networks in the first and third modules~\cite{Saini21,Mittal21b}. The fourth class of techniques embed both queries and bid phrases into the same space using a Siamese network~\cite{Huang13,Bai18,Changlarge20,Reimers19} or two-tower model~\cite{Yi19,Krichene18}, and make predictions for a novel query by retrieving the nearest embedded bid phrases with highest cosine similarity or other metrics. Section~\ref{sec:results} demonstrates that \alg could yield significant improvements in online metrics for multiple short text applications over large  ensembles of state-of-the-art methods for each class of techniques, that are currently in production in Bing.

\section{The \fm framework}\label{sec:fm}
\textbf{Notation}: Let $L$ be the number of labels and $V$ be the vocabulary size if the input is a text document. Each of the $N$ training points is then represented as $(\vx_i,\vy_i)$, where $\vx_i$ is a data-point represented  either as a dense vector, sequence of tokens or sparse bag-of-tokens depending on the application and $\vy_i \in \bc{-1,+1}^L$ is the ground truth label vector with $y_{il} = +1$ if label $l \in [L]$ is relevant to data point $i$ and $y_{il} = -1$ otherwise.

\textbf{Components}: \fm has the following components. First, a feature architecture $\cZ$ that maps a data point $\vx_i$ onto a dense $D$ dimensional representation $\hat\vx_i$, {\it i.e.,} $\cZ: \vx_i \rightarrow \hat\vx_i \in \bR^D$. Second, a surrogate objective to train intermediate feature representations. Third, a sub-linear search structure and negative sampling procedure. Fourth, a transfer mechanism to obtain final feature representations and, finally, parameters $\vW$ of a classifier model to make final predictions. Section~\ref{sec:alg} includes efficient choices made by the \alg algorithm for all these components.

\textbf{Summary}: Given a task-specific loss function $\ell$ that measures the accuracy of a classifier model on a specific label, an ideal training strategy would train $\cZ, \vW$ jointly, taking into account all positive and all negative labels of all $N$ training points, {\it i.e.,} solve $\argmin_{\cZ,\vW} \cL(\cZ,\vW)$
\[
 \cL(\cZ,\vW) = \sum_{i=1}^N\sum_{l=1}^L \ell(\vx_i, y_{il}; \cZ, \vW)
\]
However, this strategy requires jointly learning $\Om{VD + LD}$ parameters (to describe $\cZ$ and $\vW$) using the objective $\cL$ for which computing even a single gradient takes $\Om{NLD}$ time (as the objective has $NL$ terms) and is prohibitive for even moderate scale datasets. This remains true even if the objective uses a loss that does not decompose over the labels such as the F-measure, {\it etc}. To remedy this, \fm proposes a \emph{modular} strategy that effectively scales to tasks with millions of labels and data points. The core idea is to solve a much cheaper ``surrogate'' task (in \textbf{Module I}) and use this solution to identify (in \textbf{Module II}) shortlists $\hat\cN_i$ of say $\bigO{\log L}$ negative labels for each data point $i \in [N]$ ({\it i.e.,} $y_{il} = -1$ for all $l \in \hat\cN_i$) that offer a good approximation to the objective, {\it i.e.,} for all $i \in [N]$, it is the case that
\[
\sum_{l: y_{il} = -1} \ell(\vx_i, y_{il}; \cZ, \vW) \approx \sum_{l \in \hat\cN_i} \ell(\vx_i, y_{il}; \cZ, \vW).
\]
For example, if using a margin loss function such as the hinge loss, a shortlist of all margin violator negative labels would suffice. More generally, since these labels are expected to incur high loss values and also be most likely to be confused for a positive label, they are commonly referred to as \emph{hard negatives}~\cite{Jain19}. After performing a feature transfer from the surrogate to the original task (in \textbf{Module III}), final training is done (in \textbf{Module IV}) to learn $\cZ, \vW$ by solving a much less expensive optimization problem with the objective $\hat\cL$
\[
\hat\cL(\cZ,\vW) = \sum_{i=1}^N\sum_{l \in \hat\cN_i \cup \cP_i} \ell(\vx_i, y_{il}; \cZ, \vW),
\]
where for any $i \in [N]$, $\cP_i := \bc{l: y_{il} = +1}$ denotes the set of positive labels of that data point. Minimizing $\hat\cL(\cZ,\vW)$ is expected to yield parameters that resemble those obtained by minimizing $\cL(\cZ,\vW)$ due to the way hard negatives are designed. However, each module in the \fm framework can be executed in no more than $\bigO{ND\log L}$ time if careful choices are made. Finally, note that \fm supports learning an ensemble of multiple learners by training them from scratch or else simply training a re-ranker to cut down the training costs~(See section~\ref{sec:alg}).

\textbf{Flexibility}: \fm offers the flexibility of tackling a range of disparate applications with diverse inputs ranging from documents to images to graphs by letting practitioners plug in the components of their choice with minimal effort and without having to redesign the entire architecture from scratch for each application. Various feature architectures, ranging from convolutions to transformers, as well as classifier architectures, ranging from 1-vs-All to trees, can be plugged in depending on the accuracy and scalability requirements. Furthermore, \fm offers the flexibility of using diverse types of deep extreme classifiers, ranging from XML-CNN to MACH, by casting them in the proposed framework. Finally, \fm also offers the flexibility to incorporate metadata such as label features~\cite{Mittal21} or label correlations~\cite{Mittal21b} into the various modules to obtain superior performance. This allowed \fm to be used for a number of applications including text ads, product ads, rich ads, native ads, retail product recommendation, news recommendation, personalized query recommendation, {\it etc}.

\subsection{Module I: Intermediate representation}
In this module, an intermediate feature architecture $\cZ^0$ is trained using a \emph{surrogate} task. Several considerations need to be kept in mind while choosing a feature architecture and surrogate task. The feature architecture $\cZ^0$ should be trainable from available data especially  with respect to rare tokens and rare labels. $\cZ^0$ should also be able to efficiently embed data points $\vx_i \mapsto \hat\vx_i^0$, ideally in time $\bigO{cND\log L}$ where $c$ is some constant depending on the architecture, to satisfy the requirements of low-latency and high-throughput applications. Simultaneously, the surrogate task should be solvable faster than solving $\argmin_{\cZ,\vW} \cL(\cZ,\vW)$, ideally offering gradient computations in time $\bigO{cND\log L}$. It should also promote learning of a feature architecture $\cZ^0$ whose data point representations, say $\hat\vx^0_i = \cZ^0(\vx_i)$, closely resemble those offered by $\cZ$, say $\hat\vx_i = \cZ(\vx_i)$. Recall that $\cZ$ is the feature architecture that could have been learnt by directly solving $\argmin_{\cZ,\vW} \cL(\cZ,\vW)$. Doing so ensures that $\hat\vx^0_i$ and $\hat\vx_i$ have approximately the same nearest neighbors and label shortlists generated using $\hat\vx^0_i$ (in Module II) are apt proxies for those that could have been generated using $\hat\vx_i$.

The feature architecture $\cZ^0$ may be learnt in several ways. Unsupervised surrogate tasks include skip-gram models~\cite{Joulin17,Mikolov13}, next sentence prediction~\cite{Devlin19}, masked language modeling~\cite{Devlin19}, and multi-task learning~\cite{Liu19}. While scalable, architectures learnt using unsupervised training may lie far away from, and were empirically found to be 4-5\% less accurate than, those learnt using supervised training~(see Table~6 in the \suppl). Supervised learning techniques cut down the training cost by reducing the effective number of labels to $\hat{L} \ll L$. This can be implicitly done by sampling a small sub-set of labels in an online manner, say by mini-batch negative sampling~\cite{Guo19}, or explicitly by label selection, label projection or label clustering. Label selection methods~\cite{Bi13,Barezi19} select a subset of labels $\kL$ but were found to offer sub-optimal accuracies. These methods suffer from poor token coverage: if the set of data points tagged with at least one label in $\kL$, {\it i.e.,} $\bc{i : y_{ij} = +1 \text{ for any } j \in \kL}$ do not cover all vocabulary tokens, then either pre-trained token embeddings have to be externally sourced or else performance may suffer. Ensuring perfect token coverage is usually challenging -- nearly 1M labels had to be selected in order to cover 95\% of the vocabulary on the Q2B-3M dataset which defeats the very purpose of creating a surrogate task. Low-rank projection methods~\cite{Medini19,Bhatia15,Tagami17,Mineiro15} project labels on to a low-dimensional space as $\hat{\vy}_i=\vP {\vy}_i$, where $\vP \in \bR^{\hat{L}\times L}$ is a projection matrix. These are theoretically well understood but offered accuracies similar to label selection approaches in our experiments. Label clustering approaches~\cite{Prabhu18b,Khandagale19,You18,Siblini18a,Mittal21} cluster labels and treat each cluster as a \emph{meta label} using either explicit label features (using state-of-the-art encoders~\cite{Devlin19,Joulin17}) or else using indirect label representations~\cite{Jain19,Prabhu18b}. Label clustering-based approaches were found to offer the best performance in the experiments reported in section~\ref{sec:results} but other approaches might be more suitable for other applications.

\subsection{Module II: Negative sampling}
\label{sec:phase-2}
In this module, $\cZ^0$ is used to obtain intermediate representations $\hat\vx^0_i = \cZ^0(\vx_i)$ for all data points which are then used to obtain shortlists $\hat\cN_i \subset [L]$ of the $\bigO{\log L}$ most confusing or ``hardest'' negative labels, for every data point $i \in [N]$. Such deliberate hard negative mining outperforms cheaper alternatives such as mini-batch sampling~\cite{Guo19}, or sampling negatives from the power law distribution~\cite{Mikolov13} (see Fig.~2 in the \suppl). This is because the probability of choosing the most confusing negative labels is negligible when the number of labels is in the millions. Two main considerations need to be kept in mind while creating these shortlists. First, the shortlist should contain negative labels most likely to be confused for positive labels to offer concise and directed signals while training the classifiers. Second, the shortlist should be computable for every data point $i \in [N]$ in time sub-linear in $L$. Several options exist including those based on graphs~\cite{Jain19,MalkovY16}, trees~\cite{Khandagale19,Prabhu18b,Siblini18a}, clusters~\cite{Mittal21,You18} or hashing~\cite{Rawat19,Medini19,Bhatia15} that achieve sub-linear time negative sampling when working with fixed features. Note that it is possible to reuse the same data structure to shortlist $\bigO{\log L}$ labels during prediction as well. 

\subsection{Module III: Transfer learning}
In this module, the final form of the feature architecture $\cZ$ is created by adapting the intermediate architecture $\cZ^0$. A non-trivial transfer may be required since $\cZ^0$ is tuned for the surrogate task and not the original task. If this transfer involves any re-parametrization, any free parameters thus introduced are either trained separately or else fine-tuned jointly in Module-IV (see below). Three main considerations need to be kept in mind while performing this feature transfer. First, $\cZ$ should not impose significant computational overhead as compared to $\cZ^0$. Second, additional parameters introduced in $\cZ$ should be trainable from available training data, especially with respect to rare tokens, in time $\bigO{ND\log L}$. Finally, feature representations offered by $\cZ$ should not lie too far away from those offered by $\cZ^0$ so that hard negatives discovered in Module II continue to remain relevant for classifier training. Sophisticated transfer learning techniques~\cite{Pan2010,Wang17,Wei19,Saini21} could be deployed in this module and in particular, a method aiming for higher accuracies could choose to fine-tune $\cZ$ in its entirety across Modules I-IV, albeit at greater computational expense.

\subsection{Module IV: Classifier learning}
\label{sec:phase-4}
In this module, the classifier's parameters $\vW$ (and optionally, any free parameters in $\cZ$) are (jointly) learnt using an approximate objective that considers only the positive labels of a data point, {\it i.e.,} $\cP_i := \bc{l: y_{il} = +1}$ and the shortlisted negative labels $\hat\cN_i$.
	\[
	\hat\cL(\cZ,\vW) = \sum_{i=1}^N\sum_{l \in \hat\cN_i \cup \cP_i} \ell(\vx_i, y_{il}; \cZ, \vW)
	\]
Various classifiers including 1-vs-All~\cite{Babbar17,Babbar19,Prabhu18b,Jain19,Khandagale19,You18}, tree~\cite{Prabhu14,Jain16,Jernite17,Wydmuch18} or $k$-NN~\cite{Tagami17,Bhatia15} classifiers may serve as suitable choices. Note that for most commonly used loss functions, computing gradients $\nabla \hat\cL$ takes only $\bigO{ND \log L}$ time since data points typically contain only logarithmically many positive labels, {\it i.e.,} $\abs{\cP_i} \approx \bigO{\log L}$ and $\abs{\hat\cN_i} \leq \bigO{\log L}$ by design. Modules I and IV can use distinct loss functions aimed at promoting recall and precision respectively.

\section{The \alg algorithm}
\label{sec:alg}

\textbf{Module I}: \alg uses the following feature architecture that can be learnt from limited training data and be computed in 30 $\mu s$ on a CPU thereby meeting the accuracy and latency requirements of short text applications. In particular, \alg operates with sparse bag-of-words representations for documents, {\it i.e.} $\vx_i \in \bR^V$, and learns $D$ dimensional embeddings for each vocabulary token $\ve_t \in \bR^D: t \in [V]$. The intermediate features used by \alg are of the form $\hat\vx^0 := \cZ^0(\vx) = \vv + \delta\vv$ where $\vv := \text{ReLU}\br{\sum_{t=1}^{V} x_t\cdot \ve_t}$, {\it i.e.} a ReLU non-linearity over the TF-IDF weighted linear combination of the learnt token embeddings, and $\delta\vv := \text{ReLU}(\vR^0\vv)$ where $\vR^0 \in \bR^{D \times D}$ is a residual matrix. The final features are of the form $\hat\vx := \cZ(\vx) = \vv + \Delta\vv$ where $\Delta\vv := \text{ReLU}(\vR\vv)$ and $\vR \in \bR^{D \times D}$. Note that $\cZ^0(\vx)$ and $\cZ(\vx)$ share the component $\vv$ and only differ in the residual component $\delta\vv, \Delta\vv$. Restricting the spectral norms of $\vR^0$ and $\vR$ encourages a high fidelity transfer in Module III, {\it i.e.,} $\hat\vx \approx \hat\vx^0$. Thus, \alg's feature architectures $\cZ^0, \cZ$ are parametrized using token embeddings $\vE = [\ve_1,\ldots, \ve_V] \in \bR^{D \times V}$ and $\vR^0, \vR \in \bR^{D \times D}$. 

\alg adopts label clustering for its surrogate task as this was empirically observed to lead to the highest classification accuracies while keeping training time to within a few hours on a single GPU across all datasets in Table~\ref{tab:results_main}. Label centroids, defined as $\vmu^s_l= \frac{\hat\vmu^s_l}{\norm{\hat\vmu^s_l}_2}$, were used to cluster the labels, where, $ \hat\vmu^s_l = \frac1{|\cP_l|} \sum_{i \in \cP_l} \vx_i$, and $\cP_l := \bc{ i : y_{il} = +1}$ is the set of documents for which label $l$ is relevant. The balanced 2-means++ algorithm~\cite{Prabhu18b} was used to recursively cluster the labels into balanced partitions until $\hat{L}$ clusters were obtained. These clusters were treated as meta-labels and new (meta) label vectors $\hat\vy_i \in \bc{-1,+1}^{\hat L}$ were created for each training document as $\hat{y}_{ik} = +1$ for documents $i$ tagged with at least one label in cluster $k$ and $\hat{y}_{ik} = -1$ otherwise. 
Using label correlations for improved clustering led to a 2\% improvement in recall as compared to Parabel's clusters which ignored label correlations. Predicting the relevant clusters for a given document was taken as the surrogate task for Module I. 1-vs-All classifiers, parametrized as $\hat\vW = [\hat\vw_1, \ldots, \hat\vw_{\hat{L}}] \in \bR^{D \times \hat{L}}$ and $\cZ^0$ were trained jointly by solving
\begin{align}
\label{equ:surr_opt}
\argmin_{\cZ^0, \hat\vW} & \sum_{i = 1}^N \sum_{k = 1}^{\hat{L}} \log\br{1 + \exp\br{-\hat{y}_{ik}\cdot\hat\vw_k^\top\hat\vx^0_i}}
\end{align}
subject to the constraint $\sup_{\vu \neq \vzero} \norm{\vR^0\vu}_2/\norm\vu_2 \leq \lambda$, and with $\hat\vx^0_i = \cZ^0(\vx_i)$. Note that $\hat\vW, \vR^0$ are distinct from their counterparts $\vW, \vR$ used to solve the original task in Module IV. Indeed, $\hat\vW, \vR^0$ were summarily discarded after completion of Module I. However, $\vE$ was shared by $\cZ^0$ and $\cZ$ and hence, retained. The power iteration method~\cite{Miyato18} was used to constrain the spectral norm of $\vR^0$. Adding dropout after each ReLU layer was found to be as effective as using an explicit regularizer $\frac12\sum_l\hat\vw_l^\top\hat\vw_l$. Using $\hat L \leq \bigO{\log L}$, efficient parallelization on a GPU and the Adam optimizer controlled the complexity of the surrogate task. In practice, $\cZ^0$ could be trained in a couple of hours on a single GPU across all datasets in Table~\ref{tab:results_main}. 

\textbf{Module II}: The sampling cost of techniques discussed in Section~\ref{sec:phase-2} can be larger than $O(ND\log L)$ if document features $\hat\vx^0_i$ keep changing during training due to $\cZ^0$ getting updated with each mini-batch. \alg tackles this challenge by freezing the intermediate representations after completion of Module I before training a sub-linear search data structure to discover hard negatives using multiple label representations. For simplicity, \alg utilizes $\vv$ to sample hard negatives which were found to be just as effective as those sampled using $\hat{\vx}^0$. A four-step procedure was adopted to sample hard negatives. First, multiple representations were computed for each label (refer to section~A.4 in the \suppl). Then, two Approximate Nearest Neighbour Structures~(ANNS)~\cite{MalkovY16} were deployed: $\text{ANNS}^x$ over $\bc{\vv_i: i \in [N]}$ and $\text{ANNS}^\mu$ over label representations defined as $\vmu^0_l= \frac{\hat\vmu^0_l}{\norm{\hat\vmu^0_l}_2}$ where $ \hat\vmu^0_l = \frac1{|\cP_l|} \sum_{i \in \cP_l} \vv_i$. Recall that $\vv_i$ is the component shared by $\hat\vx^0_i$ and $\hat\vx_i$. The graphs were queried to generate $\bigO{\log L} \leq 500$ negative labels for each $i \in [N]$ as: (a) $\hat\cN^x_i = \bc{l : j \in \text{ANNS}^x(\vv_i), y_{jl} = +1, y_{il} = -1 }$, and (b) $\hat\cN^\mu_i = \bc{l : \vmu^0_l \in \text{ANNS}^\mu(\vv_i), y_{il} = -1}$. Adding $\bigO{\log L} \leq 50$ random negatives to the shortlist (to account for minor distortions between $\vv_i$ and $\hat\vx_i$) increased prediction accuracy.

\alg's shortlisting strategy could recall 5\% more relevant labels and the overall predictions could be 13\% more accurate than Slice. Theorem~\ref{thm:sim} indicates that it was justified to generate ANNS shortlists using $\vv_i$ rather than the final features, i.e. $\hat{\vx}_i$. This was also empirically validated as more than 87\% overlap was observed between the label shortlists computed on $\vv_i$ and those computed on $\hat{\vx}_i$. \alg's negative sampling step incurs a cost of $O(ND \log L)$, assuming $N \leq L^{\bigO1}$ and took only a few minutes on most datasets and at most 2 hours on the AmazonTitles and Q2B datasets with up to 3 million labels. Note that the shortlist could be further extended with the labels selected based on explicit label features, however label features are beyond the scope of \alg.

\begin{theorem}
\label{thm:sim}
Let $\lambda$ be the spectral norm of $\vR$, $\cP_l = \{i | y_{il} = +1\}$ be the set of positive training points for the label $l$, $\epsilon_l = (1+\lambda\sqrt{|\cP_l|})^2-1$, and $\vmu_l$ be the label representation that could have been computed using final features as $\vmu_l= \frac{\hat\vmu_l}{\norm{\hat\vmu_l}_2}$ where $ \hat\vmu_l = \frac1{|\cP_l|} \sum_{i \in \cP_l} \hat\vx_i$ and $\hat\vx_i = \cZ(\vx)$. Then: (a) $\norm{\hat\vx_i - \vv_i}_2 \leq \lambda\norm{\vv_i}_2 \; \forall i$, i.e. the final and intermediate features for any document can be brought arbitrarily close by restricting $\lambda$; and (b) $\frac{C(\vv_i, \vmu^0_l)}{1+\epsilon_l} \leq C(\hat\vx_i, \vmu_l) \leq C(\vv_i, \vmu^0_l) + \epsilon_l$, i.e. the cosine similarity $C(\va, \vb) := \frac{\va^\top \vb}{\norm\va_2\norm\vb_2}$ used in the ANNS algorithm to determine the similarity between any document $i$ and label $l$ can also be made arbitrarily close between the intermediate and final representations by restricting $\lambda$, the spectral norm of $\vR$. [Please refer to~A.6 in the \suppl for a proof.]
\end{theorem}

\textbf{Module III}: \alg uses a final feature representation of the form $\hat\vx : = \cZ(\vx) = \vv + \Delta\vv$, where $\vv$ is shared with $\cZ^0$ and $\Delta\vv = \text{ReLU}(\vR\vv)$ for a residual matrix $\vR$. Note that $\vR$ is distinct from the residual matrix $\vR^0$  used in $\cZ^0$. Since $D^2 \ll VD$, this introduced a negligible increase in model parameters and computational complexity and $\Delta\vv$ could be computed in under 10 $\mu s$ on a CPU thereby meeting the low-latency constraints. The spectral norm of $\vR$ was restricted to be no greater than $\lambda \leq 1$ to ensure that feature representations offered by $\cZ$ did not lie far away from those offered by $\cZ^0$ (see Theorem~\ref{thm:sim}). Prediction accuracy was observed to degrade if the spectral norm was not restricted or if the residual block was replaced by a fully connected block~\cite{Medini19} since such steps allowed the final features to drift away from the intermediate features.

\textbf{Module IV}: \alg adopts the high capacity 1-vs-All classifier model and learns a $D$-dimensional classifier per label. Thus, its classifier is parametrized as $\vW = [\vw_1,\ldots, \vw_L] \in \bR^{D \times L}$. The classifier and residual block $\vR$ present in $\cZ$ were jointly trained in $O(ND \log L)$ time by restricting training to the positive and shortlisted negative labels for each data point. Specifically, for any $i \in [N]$, let $\cP_i := \bc{l: y_{il} = +1}$ be the set of positive labels of the data point, $\hat\cN_i = \hat\cN^\mu_i \cup \hat\cN^z_i$ be the negative labels shortlisted in Module II and let $\hat\cS_i := \hat\cN_i \cup \cP_i$. Then the approximate objective $\argmin_{\vR, \vW}\hat\cL(\vR,\vW)$ was solved where
\[
\hat\cL(\vR,\vW) = \sum_{i=1}^N\sum_{l \in \hat\cS_i} \log\br{1+\exp\br{-y_{il}\cdot\vw_l^\top\hat\vx_i}},
\]
subject to the constraint $\sup_{\vu \neq \vzero} \norm{\vR\vu}_2/\norm\vu_2 \leq \lambda$, where $\vW = [\vw_1,\ldots, \vw_L] \in \bR^{D \times L}$ are the 1-vs-All classifiers and $\hat\vx_i = \cZ(\vx_i)$. This formulation could be optimized efficiently using the Adam optimizer on a single P40 GPU in 0.25--5 hours on the datasets considered in this paper.

\textbf{Re-ranking the labels}: \alg improves its accuracy by learning a novel re-ranker as follows. \alg's predictions $ \hat{\vy}_i \in \bc{-1,+1}^L$ were obtained for each training point $i$. Excluding the true positives yielded a shortlist of negative labels that \alg found the most confusing for each point. A re-ranker was then trained to eliminate these mis-predictions by optimizing $\argmin_{\widetilde\cZ, \widetilde\vW}\tilde\cL(\widetilde\cZ, \widetilde\vW)$ where
\[
\tilde\cL(\widetilde\cZ, \widetilde\vW) = \sum_{i=1}^N \sum_{l \in \widetilde\cS_i} \log\br{1 + \exp\br{-y_{il}\cdot\widetilde\vw_l^\top\widetilde\vx_i}},
\]
where $\widetilde\cS_i = \bc{l : y_{il} = +1} \cup \bc{l : \hat{y}_{il} = +1}$, $\widetilde\vx_i = \widetilde\cZ(\vx_i)$ and $\widetilde\cZ$ had an architecture similar to $\cZ$ but with independent parameters $\widetilde\vE, \widetilde\vR$. This increased accuracy by $1$--$3\%$ with comparatively larger gains on larger datasets, with only a 10--20\% increase in training time. 

\textbf{Log-time prediction}: \alg meets the latency requirements of online short text applications on even the largest datasets by making predictions in $O(D^2 + D \log L)$ time. Given a test document $\vx \in \bR^V$, the $O(\log L)$ most relevant labels $\hat\cS :=  \cN^\mu \cup \cN^x$ along with their base similarity scores $s^l(\vv)$ were shortlisted using $\vv := \text{ReLU}\br{\sum_{t=1}^{V} x_t\cdot \ve_t} \in \bR^D$. Next, the final features $\hat\vx = \cZ(\vx)$ and 1-vs-All classifiers were used to yield scores $\hat{y}_l = \alpha\sigma(\vw_l^\top\hat\vx) + (1-\alpha)\sigma(s^l(\vv))$ for shortlisted labels $l \in \hat\cS$ and $\hat{y}_l = 0$ otherwise where $\sigma$ is the sigmoid function and $\alpha \in [0, 1]$ is a hyperparameter. Final predictions were given by linearly combining the re-ranker scores $\tilde{y}_l = \sigma( \widetilde\vw_l^\top\widetilde\vx)$ with the base similarity scores as $\overline{y}_l = \beta \hat{y}_l + (1-\beta)\widetilde{y}_l$ if $l \in \hat\cS$ and $\overline{y}_l = 0$ otherwise, using a hyper-parameter $\beta \in [0,1]$.

\begin{table}
    \caption{\alg could be significantly more accurate and scalable than leading deep extreme classifiers including MACH, XML-CNN and AttentionXML on publicly available benchmark datasets. \alg was also found to outperform leading methods designed to match user queries to bid phrases on the Q2B-3M dataset. Results for other methods and metrics are presented in the \suppl.}
    \label{tab:results_main}
      \centering
      \resizebox{\linewidth}{!}{
        \begin{tabular}{@{}l|ccc|cc|c@{}}
        \toprule
        \textbf{Method} & \textbf{P@1} & \textbf{P@3} & \textbf{P@5} & \textbf{PSP@3} & \textbf{PSP@5} &  \multicolumn{1}{c}{\begin{tabular}[c]{@{}c@{}}\textbf{Training}\\ \textbf{Time (hr)}\end{tabular}} \\
        
        \midrule
        
        \multicolumn{7}{c}{Q2B-3M}\\ \midrule						
        \alg	 & \textbf{73.37}	 & \textbf{33.91}	 & \textbf{21.67}	 & \textbf{85.13} & \textbf{90.42} & 15.59 \\
        Parabel	 & 54.29	 & 27.15	 & 17.94	 & 61.66 & 68.61 & 3.52\\
        Slice+CDSSM	 & 53.23	 & 27.53	 & 18.56 & 64.51 & 74.14 & 4.71\\
        Seq2Seq	 & 28.25	 & 13.06	 & 8.02	 & 17.59 & 15.42 & -\\
        Simrank++	 & 52.70	 & 29.69	 & 19.33	 & 41.67 & 39.36 & 25.00\\
        
        \midrule
        
        \multicolumn{7}{c}{AmazonTitles-670K}\\ \midrule						
        \alg	 & 39.97	 & 35.73	 & 32.59	 & 29.79 & 31.71 & 1.29 \\
        \alg-3	 & \textbf{40.63}	 & \textbf{36.22}	 & \textbf{33.00}	 & \textbf{30.17} & \textbf{32.07} & 3.85\\
        MACH	 & 34.92	 & 31.18	 & 28.56	 & 23.14 & 25.79 & 6.41\\
        XML-CNN	 & 35.02	 & 31.37	 & 28.45     & 24.93 & 26.84 & 23.52\\
        Slice+fastText	 & 33.85	 & 30.07	 & 26.97	 & 24.15 & 25.81 & 0.22\\
        AttentionXML	& 37.92	 & 33.73	 & 30.57	 & 26.43 & 28.39 & 37.50\\
        Parabel	& 38.00	 & 33.54	 & 30.10	 & 25.57 & 27.61 & 0.09\\
        Bonsai	 & 38.46	 & 33.91	 & 30.53	 & 26.19 & 28.41 & 0.53\\
        DiSMEC	 & 38.12	 & 34.03	 & 31.15	 & 25.46 & 28.67 & 11.74\\

        \midrule
        
        \multicolumn{7}{c}{AmazonTitles-3M}\\ \midrule						
        \alg	 & 47.64	 & 44.66	 & 42.36	 & 18.59 & 20.60 & 4.38 \\
        \alg-3	 & \textbf{48.74}	 & \textbf{45.70}	 & \textbf{43.31}	 & \textbf{18.89} & \textbf{20.94} & 13.04\\
        MACH	 & 37.10	 & 33.57	 & 31.33	 & 8.61 & 9.46 & 40.48\\
        SLICE+fastText	 & 35.39	 & 33.33	 & 31.74	 & 13.37 & 14.94 & 0.64\\
        Parabel	& 46.42	 & 43.81	 & 41.71	 & 15.58 & 17.55 & 1.54\\
        Bonsai	 & 46.89	 & 44.38	 & 42.30	 & 16.66 & 18.75 & 9.90\\
        \midrule
        
        \multicolumn{7}{c}{WikiSeeAlsoTitles-350K}\\ \midrule						
        \alg	 & 20.42	 & 14.44	 & 11.39	 & 12.05 & 13.94 & 1.47 \\
        \alg-3	 & \textbf{20.61}	 & \textbf{14.58}	 & \textbf{11.49}	 & \textbf{12.16} & \textbf{14.04} & 4.36\\
        MACH	 & 14.79	 & 9.57	 & 7.13	 & 7.02 & 7.54 & 7.44\\
        XML-CNN	 & 17.75	 & 12.34	 & 9.73 & 9.72 & 11.15 & 14.25\\
        Slice+fastText	 & 18.13	 & 12.87	 & 10.29	 & 10.78 & 12.74 & 0.22\\
        AttentionXML	& 15.86	 & 10.43	 & 8.01	 & 7.20 & 8.15 & 30.44\\
        Parabel	& 17.24	 & 11.61	 & 8.92	 & 8.83 & 9.96 & 0.06\\
        Bonsai	 & 17.95	 & 12.27	 & 9.56	 & 9.68 & 11.07 & 0.46\\
        DiSMEC	 & 16.61	 & 11.57	 & 9.14	 & 9.19 & 10.74 & 6.62\\
        
        \midrule
        
        \multicolumn{7}{c}{WikiTitles-500K}\\ \midrule						
        \alg	 & 46.01	 & 25.62	 & 18.18	 & 18.59 & 18.95 & 4.45 \\
        \alg-3	 & \textbf{46.60}	 & \textbf{26.03}	 & \textbf{18.50}	 & \textbf{18.90} & \textbf{19.30} & 13.04\\
        MACH	 & 33.74	 & 15.62	 & 10.41	 & 8.98 & 8.35 & 23.65\\
        XML-CNN	 & 43.45	 & 23.24	 & 16.53 & 14.74 & 14.98 & 55.21\\
        Slice+fastText	 & 28.07	 & 16.78	 & 12.28	 & 14.69 & 15.33 & 0.54\\
        AttentionXML	& 42.89	 & 22.71	 & 15.89	 & 14.32 & 14.22 & 102.43\\
        Parabel	& 42.50	 & 23.04	 & 16.21	 & 16.12 & 16.16 & 0.34\\
        Bonsai	 & 42.60	 & 23.08	 & 16.25	 & 16.85 & 16.90 & 2.94\\
        DiSMEC	 & 39.89	 & 21.23	 & 14.96	 & 15.15 & 15.43 & 23.94\\
        \bottomrule
    \end{tabular}}
\end{table}
\section{Experiments} \label{sec:results}

\textbf{Datasets}: Results are presented on publically available benchmark short text datasets with up to 3 million labels for predicting frequently bought together Amazon items based on just the product title (AmazonTitles-670K and AmazonTitles-3M) as well as using a Wikipedia article's title to predict it's Wikipedia tags (WikiTitles-500K) as well as its related Wikipedia articles (WikiSeeAlsoTitles-350K). All datasets are available on the Extreme Classification Repository~\cite{XMLRepo}. Note that, even though the focus of this paper is on short text applications and \alg has been designed keeping their specific requirements in mind, results on benchmark long text datasets are presented in the \suppl for completeness. Results are also presented on a proprietary Bing dataset with 3 million labels and 21 million training points for matching user queries to advertiser bid phrases (Q2B-3M). The dataset was created by mining Bing's click logs where each user's query was treated as a data point and clicked advertiser bid phrases became its labels. Table~3 in the \suppl presents the data set statistics.
 
 \textbf{XML baselines}: The focus of this paper is on comparing \alg to MACH~\cite{Medini19} and Slice~\cite{Jain19} as they have been specifically designed for short text applications. \alg was also compared to other leading deep extreme classifiers including XML-CNN~\cite{Liu17}, XTransformers~\cite{Chang20} and AttentionXML~\cite{You18}. Furthermore, for the sake of completeness, results are presented for non-deep extreme classifiers including XT~\cite{Wydmuch18}, DiSMEC~\cite{Babbar17}, PfastreXML~\cite{Jain16}, Parabel~\cite{Prabhu18b}, Bonsai~\cite{Khandagale19} and AnneXML~\cite{Tagami17}. Implementations of all the baseline algorithms were provided by their authors. The hyper-parameters of these algorithms were set as suggested by their authors wherever applicable and by fine-grained validation otherwise. Results are only presented for those datasets to which an implementation could scale. Results could therefore not be reported for XTransformers  as it could not be trained on any of the datasets on a single GPU in a week. As is customary in extreme classification, results are presented for not only \alg but also an ensemble with 3 learners referred to as \alg-3. \alg's hyper-parameters and their settings on various datasets are discussed in the \suppl.

\textbf{Bing baselines}:
The online flight results revealed the gains that \alg was able to achieve when added to a large ensemble of state-of-the-art techniques currently running in production on Bing including many leading techniques for query synthesis (constrained and unconstrained), graph based techniques (graph neural networks, random walks, sessions based methods, {\it etc.}), embedding methods (Siamese networks and two-tower models), extreme classifiers as well as techniques that leverage additional information (refer to Section~\ref{sec:lit}). Offline results for some of these techniques such as Simrank++~\cite{Ioannis08} and a BERT based sequence-to-sequence constrained synthesis model~\cite{Lian19, Devlin19} are presented on the Q2B-3M dataset for matching queries to bid phrases.

\textbf{Evaluation metrics}: Performance was evaluated using prec-ision@$k$ (P@$k$), and propensity scored precision@$k$ (PSP@$k$) which have been widely used in the extreme classification literature~\cite{Babbar17,Jain16}. Results on additional metrics such as nDCG@$k$ (N@$k$) and propensity scored nDCG@$k$ (PSN@$k$) have been included in \suppl which also contains the definitions of all the metrics. All training times have been reported on a 24-core Intel Xeon 2.6 GHz machine with a single Nvidia P40 GPU unless stated otherwise.

\textbf{Table~\ref{tab:results_main} - Offline results}: \alg's main competitors were Slice and MACH as they are extreme classifiers that have been developed for low-latency short text applications since their features could be extracted in milliseconds on a CPU while all other architectures required a GPU. Slice is not a deep learning method and was therefore trained on the same FastText embeddings that were used to initialize \alg. Nevertheless, \alg was 2.29-17.94\% more accurate than Slice and 5.05-12.27\% more accurate  than MACH on the publically available Repository datasets and 20.14\% more accurate than Slice on the Bing Q2B-3M dataset. Unfortunately, MACH could not be trained on the Q2B-3M dataset in a week on a single GPU whereas \alg could be trained in 13 hours. On the Repository datasets, \alg was 5-9$\times$ faster to train than MACH. Similarly, none of the other deep extreme classifiers could be trained on the AmazonTitles-3M and Q2B-3M datasets. On the smaller datasets, \alg was 2.05-4.56\% more accurate and 9-29$\times$ faster to train than AttentionXML and XML-CNN respectively. Furthermore, \alg was at least 19\% more accurate than all other methods on the Bing Q2B-3M dataset including SimRank++ and a BERT based sequence-to-sequence model that had been specifically designed for matching queries to bid phrases. These results demonstrate that \alg's features could be learnt accurately from limited training data and that the \fm framework enabled \alg to be significantly more scalable than all other deep extreme classifiers. Finally, for the sake of completeness, \alg was compared to non-deep learning extreme classifiers on the Repository datasets where it could be up to 6.12\% more accurate and this increased marginally to 6.77\% for the \alg-3 ensemble.

\textbf{Online results from Bing flights}: \alg was able to efficiently train in 20 hours on 4$\times$P40 GPUs on various Bing internal datasets with up to 62 million labels that were far beyond the scaling capabilities of all other deep extreme classifiers. Furthermore, \alg's features could be extracted in microseconds on a CPU and its overall predictions made in a few milliseconds allowing it to make billions of predictions per day at peak rates of 120,000 queries per second on commodity hardware. This allowed \alg to be flighted for multiple short text applications on Bing with extremely low-latencies and high-throughputs including text ads, product ads, rich ads, native ads, retail product recommendation, news recommendation, personalized query recommendation, {\it etc}. Unfortunately, due to space constraints, online results when \alg was added to the ensemble of state-of-the-art techniques currently in production can only be presented for just two of these applications.

\textbf{Personalized ads}: Billions of users were shown personalized ads as they surfed the web based on their browsing history in the first application. Each user's intent was represented by the set of queries that the user could have potentially asked on Bing to reach the last $K$ webpages that they had browsed. \alg was used to predict the set of Bing queries from each visited webpage's title in milliseconds. A GRU was used to model the user's last $K$ states in near real time. The GRU was then used to select a single predicted query that was passed through the Bing pipeline to show ads to the user. Human expert evaluation revealed that \alg increased the number of excellent predictions by more than 20\% while reducing the number of fair and poor predictions by more than 10\% as compared to the ensemble in production containing leading extreme classification, deep learning and IR techniques. \alg was also able to achieve 100\% coverage when deployed online by making predictions for all visited webpages for all users within the latency and throughput constraints. This allowed \alg to increase the click-through-rate by 6.5\% and revenue by more than 5\%. Table~8 in the \suppl shows examples comparing \alg's predictions to those of traditional approaches and demonstrates \alg's benefits over the state-of-the-art.

\textbf{Matching queries to bid phrases}: \alg treated each user's query as input and each advertiser bid phrase as a separate label in order to predict the set of bid phrases that could be matched to the user's query to show ads on Bing. \alg's offline accuracy was 19\% higher than that of other approaches~(see Table~\ref{tab:results_main}). This translated into an increase of 1.6\% in revenue per thousand queries, 2.9\% in match quality and 8.6\% increase in query coverage over a large ensemble of state-of-the-art techniques when deployed online. Table~\ref{tab:qual_analysis_q2b} lists \alg's predicted bid phrases for the query ``what is diabetes type''. As can be seen, \alg's predictions could be more accurate and diverse than those of other methods which could fixate on the phrase ``what is'' and thereby make many mispredictions.

\begin{table}
    \small
	\caption{\alg's predicted bid phrases for the user query "what is diabetes type 2" are more accurate and diverse as compared to leading methods~(M1--M3) in Bing. All mispredictions have been \textit{italicized}.}
	\label{tab:qual_analysis_q2b}
      \centering
        \begin{tabular}{@{}ll}
        \toprule
        \textbf{Method} & \textbf{Predictions} \\
        \midrule
        \alg & definition diabetes type 2, what causes type 2 diabetes, \\
        & do i have type 2 diabetes, what is type 2 diabetes mellitus\\
        & what are the causes of diabetes type 2, type 2 diabetes \\
        \midrule
        M1 & what is type ii diabetes, whats type 2 diabetes\\
        \midrule
        M2 & type 2 diabetes\\
        \midrule
        M3 & what is type 2 diabetes, what is type 1 and type 2 diabetes, \\
        & type 2 diabetes, \textit{what is email marketing}, \textit{what is ptsd2} \\
        & \textit{what is anemia}, \textit{what is radiation therapy} \\
        \bottomrule
        \end{tabular}
\end{table}

\textbf{Ablations}: Table~6 in the \suppl presents the ablation results for each module validating \alg's design choices. First, much progress has been made in both designing and speeding up feature architectures such as BERT~\cite{Devlin19}, Roberta~\cite{Liu19roberta}, {\it etc.} that can easily be incorporated into the \fm framework, if desired. Unfortunately, doing so did not lead to accuracy gains on the short text extreme datasets and \alg's features were demonstrated to be at least 2\% more accurate than those based on CNNs~\cite{Liu17}, MLPs~\cite{Medini19} and BERT~\cite{Devlin19}. This indicates that \alg's features were more suitable for short text extreme classification and that more research is required on how to train and fine-tune transformer and other architectures when many features and labels have very limited training data. Second, training on \alg's surrogate task was demonstrated to be 1-4\% more accurate than training using self-supervision, label selection or label low-rank projection indicating the suitability of label clustering and the importance of a well designed surrogate task. Third, it was demonstrated that the negative sampling strategy proposed in \alg could be 13\% more accurate than the strategy in Slice. This further highlights the differences between \alg and Slice and indicates that, even if Slice could somehow have been trained jointly along with \alg's features, it would still have been significantly inferior to \alg. Finally, it was demonstrated that \alg's performance could not be further improved by replacing its classifier with alternatives such as DiSMEC or Parabel.

\textbf{State-of-the-art algorithms in the \fm framework}: \fm could be used to analyze and improve leading deep extreme classifiers thereby demonstrating its generality and usefulness. For instance, both XML-CNN and MACH could be recast into the \fm framework by replacing \alg's feature architecture by their CNN or MLP features. This led to accuracy gains of 1.8\% and 2.4\% and training speedups of 10$\times$ and 5$\times$ for XML-CNN and MACH respectively (see Table~4 in the \suppl). Note that the original MACH algorithm stops training after the first \fm module, {\it i.e.,} after it has learnt its MLP features on the surrogate task, and therefore compensates by learning a large ensemble. Casting MACH into \fm allowed it to improve accuracy by fine-tuning its features for the extreme task at hand while speeding up training by learning just a single learner rather than an ensemble of 32 learners. Similarly, \fm increased XML-CNN's accuracy by fine-tuning its CNN features through the residual block in the third module and speeded up training by replacing its fully connected output layer with $O(L)$ complexity by a 1-vs-some classifier with $O(\log L)$ complexity in the fourth module. It should be reiterated that \alg's performance continued to remain superior to that of the improved MACH and XML-CNN. 

\textbf{Vision task}: While the primary focus of this paper is on short text applications, \fm was also applied to a vision task to demonstrate its ease-of-use and generality. Table~9 in the \suppl presents results for classifying a retail item's image into its product categories from the AmazonCat-13K dataset available on the Repository. \fm made only a minor modification in \alg by replacing its feature architecture by ResNet18~\citep{He16} while leaving everything else unchanged. Pre-trained ResNet18 features were also used to train Slice, MACH, Parabel and DiSMEC. Nevertheless, \fm was 4-29\% more accurate than these methods thereby demonstrating its flexibility.

\section{Conclusions}
This paper developed the modular \fm framework that was used to: (a) derive the \alg algorithm for low-latency short text applications; (b) analyse and improve leading deep extreme classifiers; and (c) provide a convenient and flexible tool for practitioners to plug in components of their choice with minimal effort for tackling diverse applications. Furthermore, \alg was demonstrated to be significantly more accurate and scalable than leading extreme classifiers for short text documents and could lead to significant gains in various online metrics for multiple applications on Bing.

\begin{acks}
The authors thank Purushottam Kar, Aditya Kusupati, Harsha Vardhan Simhadri, Yash Garg, and Risi Thonangi for helpful feedback.
\end{acks}

\bibliographystyle{ACM-Reference-Format}
\bibliography{references}
\clearpage
\appendix
\onecolumn
\section{Supplementary Material}
\begin{table*}
    \tiny
	\caption{Dataset Statistics. Please note that in Q2B-3M dataset, character 3-grams and 4-grams tokens were also included in the vocabulary for \alg, Parabel {\it etc.}}
	\label{tab:stats}
	\resizebox{\linewidth}{!}
	{
		\begin{tabular}{l|ccccccc}
			\toprule
			\textbf{Dataset} &
			\textbf{\begin{tabular}[c]{@{}c@{}}Train Instances \\ $N$ \end{tabular}} &
			\textbf{\begin{tabular}[c]{@{}c@{}}Features\\ $V$ \end{tabular}}  &
			\textbf{\begin{tabular}[c]{@{}c@{}}Labels\\ $L$ \end{tabular}} &
			\textbf{\begin{tabular}[c]{@{}c@{}}Number of\\ Test Instances\end{tabular}} &
			\textbf{\begin{tabular}[c]{@{}c@{}}Average Labels\\ per sample \end{tabular}} &
			\textbf{\begin{tabular}[c]{@{}c@{}}Average Points\\ per label \end{tabular}} &
			\textbf{\begin{tabular}[c]{@{}c@{}}Average Features\\ per instance \end{tabular}} \\
			\midrule
			WikiSeeAlsoTitles-350K & 629,418 & 91,414 & 352,072 & 162,491 & 2.33 & 5.24 & 2.73 \\
        WikiTitles-500K & 1,699,722 & 185,479 & 501,070 & 722,678 & 4.89 & 23.62 & 2.73 \\
        AmazonTitles-670K & 485,176 & 66,666 & 670,091 & 150,875 & 5.39 & 5.11 & 5.26 \\
        AmazonTitles-3M & 1,517,620 & 165,431 & 2,812,281 & 655,479 & 35.06 & 27.09 & 7.58 \\
        Q2B-3M & 21,561,529 & 1,284,191 & 3,192,113 & 6,995,038 & - & - & - \\
			\bottomrule
		\end{tabular}
	}
\end{table*}
 \begin{table}
    \caption{Accuracy gain ($\Delta$ \textbf{P@1}) and training speedup for leading methods in \fm framework relative to the original algorithms on the AmazonTitles-670K dataset.}
	\label{tab:deepxml_framework}
    \begin{tabular}{@{}l|c|c@{}}
    \toprule
    \textbf{Method} & $\Delta$ \textbf{P@1} & \textbf{Speed-up}\\
     \midrule
     \fm + XML-CNN & +1.89 & 10$\times$ \\
     \fm + MACH & +2.41 & 5$\times$ \\
     \bottomrule
    \end{tabular}
\end{table}

\subsection{Hyper-parameters}\label{sec:supp:hyper-parameters}
\alg's hyper-parameters include $\alpha$ for combining the classifier \& shortlist scores, $\hat{L}$ which was the number of labels in the surrogate task which was set to $2^{16}$ across all datasets and the label shortlist size which was set to $500$ in all cases. Note that the different values of $\alpha$ leads to a trade-off in vanilla and propensity scored metrics. The embeddings in $\mathcal{Z}$, residual matrices and classifiers were initialized with FastText~\cite{Joulin17}, the identity matrix and Xavier's method respectively. \alg was trained using the Adam optimizer with spectral norm constraints~\cite{Miyato18} and its hyper-parameters included the learning rate, the batch size and the number of epochs. Most of these were set to default values across datasets and the most expensive hyper-parameter to tune was the learning rate on the surrogate task. 

 Experiments were performed on a P40 GPU card with CUDA 11, and Pytorch 1.8 unless stated otherwise. Dropout with probability 0.5 was used for all datasets. HNSW~\citep{MalkovY16} parameters $M$, $efC$ and $efS$ where set to $100, 300, 300$ and $50, 50, 100$ for $\text{ANNS}^x$ and $\text{ANNS}^{\mu}$ respectively. The surrogate learning task was trained with $|\hat{L}| = 2^{16}$ with learning rate chosen from $\{0.003, 0.005, 0.02 \}$. Increasing $|\hat{L}|$ beyond $2^{16}$ lead to only marginal gains in accuracy but at the cost of increase in training time. The model parameters for the extreme task were learnt with a learning rate chosen from $\{0.002, 0.0005\}$. It should be noted that no hyper-parameter tuning was done for proprietary datasets where Astec lead to significant gains in offline as well as online metrics.

\begin{table}[ht]
    \caption{Parameter settings for \alg on different datasets. }
	\label{tab:hyperparameters}
    \begin{tabular}{l|ccc}
            \toprule
            \textbf{Dataset} & $\mathbf{|\hat{L}|}$ & \textbf{\begin{tabular}[c]{@{}c@{}}Learning Rate\\ (Surrogate task)\end{tabular}} &    \textbf{\begin{tabular}[c]{@{}c@{}}Learning Rate\\ (Extreme task)\end{tabular}}  \\ \midrule
            WikiSeeAlsoTitles-350K     & $2^{16}$ & 0.005 & 0.002 \\
            WikiTitles-500K            & $2^{16}$ & 0.005  & 0.0005 \\
            AmazonTitles-670K          & $2^{16}$ & 0.02 & 0.002 \\
            AmazonTitles-3M            & $2^{16}$ & 0.003 & 0.0005 \\
            Q2B-3M                     & $2^{16}$ & 0.02 & 0.002 \\
            \bottomrule
        \end{tabular}
\end{table}

\begin{table}
\caption{\alg's results for different choices of modules. Note that results are reported without re-ranker component and only one component was varied at a time. AmazonTitles-670K \& WikiTitles-500K were used for (a)--(c) \& (d) respectively.}
\small
\label{tab:deepxml_ablation}
\label{tab:ablation_framework}
    \begin{subtable}{0.47\linewidth}
        \caption{\small{Intermediate representation}}
        \label{tab:ablation_features}
        \centering
        \begin{tabular}{@{}l|cc}
            \toprule
            Method & \textbf{P@1}  & \textbf{P@5} \\
            \midrule
            CNN~\citep{Liu17} & 36.91 & 30.76\\
            MLP~\citep{Medini19} & 37.41 & 30.65\\
            Bert~\citep{Devlin19} & 36.15 & 29.65 \\
            \alg & 39.12 & 32.07 \\
            \bottomrule
        \end{tabular}
    \end{subtable}~~~
    \begin{subtable}{0.47\linewidth}
        \caption{\small{Surrogate task}}
        \label{tab:ablation_surrogate}
        \centering
        \begin{tabular}{@{}l|cc}
            \toprule
            Method & \textbf{P@1}  & \textbf{P@5} \\
            \midrule
            Unsupervised~\citep{Joulin17} & 34.80 & 27.42 \\
            Random~\citep{Medini19} & 38.11 & 30.97\\
            Label selection & 38.3 & 31.25 \\ 
            \alg & 39.12 & 32.07 \\
            \bottomrule
            \end{tabular}
    \end{subtable}
    
    \begin{subtable}{0.47\linewidth}
        \caption{\small{Classifier}}
    \label{tab:ablation_classifier}
    \centering
    \begin{tabular}{@{}l|cc}
        \toprule
        Classifier & \textbf{P@1}  & \textbf{P@5} \\
        \midrule
        Parabel & 37.07 & 28.75 \\
        Slice & 36.73 & 30.07 \\
        DiSMEC & 38.42 & 31.44 \\
        \alg & 39.12 & 32.07 \\
\bottomrule
    \end{tabular}
    \end{subtable}~~~~ 
    \begin{subtable}{0.47\linewidth}
        \caption{\small{Negative sampling}}
        \label{tab:ablation_ns}
        \centering
            \begin{tabular}{@{}l|cc}
                \toprule
                Method & \textbf{P@1}  & \textbf{P@5} \\
                \midrule
                Uniform & 27.3 & 11.4 \\
                NEG~\citep{Mikolov13} & 30.62 & 11.87 \\
                Slice~\citep{Jain19} & 32.33 & 14.37 \\
                \alg & 45.45 & 17.64 \\
    \bottomrule
        \end{tabular}
    \end{subtable}
\end{table}

\subsection{Clustering labels}\label{sup:clustering}
\alg clustered the labels using label centroid representation $\hat{\vmu}_l = \frac{\sum_{i=1}^{N}y_{il}\vx_i}{\norm{\sum_{i=1}^{N}y_{il}\vx_i}_2}$ as the label meta-data was unavailable. 2-means++ algorithm was deployed to solve the following optimization problem, which recursively clusters the labels into two balanced-partitions to finally end up with $\hat{L}$ clusters.

\begin{align*}\label{equ:clustering}
\argmin_{ {\vmu}_{\pm},\valpha \in \{-1, 1\}^{L}} & \sum_{l=1}^{L} \cC_{ll} \left(\frac{1+\valpha_l}{2} \hat{\vu}_l \vmu_{+} + \frac{1-\valpha_l}{2} \hat{\vu}_l \vmu_{-} \right)\\ 
&+ \sum_{l=1}^{L}\sum_{p=1}^{L} \cC_{lp} \left(\frac{1+\valpha_l}{2} \hat{\vu}_p \vmu_{+} + \frac{1-\valpha_l}{2} \hat{\vmu}_p \vmu_{-}\right) \\
\mbox{s. t.} ~~&~~ \norm{\vmu_{\pm}}_2 = 1,~~-1 \le \sum_{l=1}^{L} \valpha_l \le 1
\end{align*}

where it has been assumed without loss of generality that $L$ labels need to be partitioned at each step, $\alpha_l=\pm1$ means that label $l$ is assigned to cluster with mean $\vmu_{\pm}$ and $\cC$ is the $\ell1$ normalized label correlation matrix~($\vY^{\top} \vY$). In practise, the label correlation matrix was estimated by performing a random walk over a graph with labels as nodes and data-points as edges.

\subsection{Additional surrogate tasks} 
\label{subsec:divalgo}
A simple but effective way to train the parameters of chosen feature architecture, {\it i.e.,} $\mathcal{Z}$ could be to select $\hat{L}$ labels based on label frequency in the training set. The hyper-parameter $\hat{L}$ was chosen to balance two constraints. First, $\hat{L}$ should be large enough so that almost all the token embeddings could be learnt in this first phase of training. At the same time, the $|\hat{L}|$ should be small enough so that (\ref{equ:surr_opt}) could be optimized efficiently on a single GPU as a non-extreme problem and without resorting to ANNS shortlisting. It was empirically observed that setting $0.05L \leq \hat{L} \leq 0.2L$ with lower values being preferred for larger problems resulted in accurate intermediate representations. It should be reiterated that the balanced clustering was found to be more accurate and scalable than the alternatives including label selection based techniques.

\subsection{ANNS search and multiple representatives}\label{sup:multiple_rep}
It is worth pointing out that some of the most frequently occurring head labels could be usefully represented by {\em multiple} $k$-means cluster centres while constructing the $\mbox{ANNS}$ small world graph. This allowed for the accurate shortlisting of multi-modal head labels which could not be shortlisted well based on a single label centroid representation, {\it i.e.,} $\vmu_l^0$. For instance, representing the top 4 head labels on the WikiTitles-500K dataset by 300 $k$-means cluster centres rather than just the label mean improved recall@300 by $5\%$ without any noticeable increase in the training or prediction time.

\subsection{Evaluation metrics}
\label{sup:eval}
Performance has been evaluated using propensity scored precision@$k$ and nDCG@$k$, which are unbiased and more suitable metric in the extreme multi-labels setting~\citep{Jain16, Babbar19, Prabhu18, Prabhu18b}. The propensity model and values available on The Extreme Classification Repository~\citep{XMLRepo} were used. Performance has also been evaluated using vanilla precision@$k$ and nDCG@$k$ (with $k$ = 1, 3 and 5) for extreme classification. 

For a predicted score vector $\hat{\mathbf{y}} \in R^L$ and ground truth vector $\mathbf{y} \in \{0, 1\}^L$:
		$$P@k = \frac{1}{k} {\sum_{l \in rank_k(\hat{\mathbf{y}})}} y_l$$
		$$PSP@k = \frac{1}{k} {\sum_{l \in rank_k(\hat{\mathbf{y}})}} \frac{y_l}{p_l}$$
		$$DCG@k = \frac{1}{k} {\sum_{l \in rank_k(\hat{\mathbf{y}})}} \frac{y_l}{\log(l+1)}$$
		$$PSDCG@k = \frac{1}{k} {\sum_{l \in rank_k(\hat{\mathbf{y}})}} \frac{y_l}{p_l \log(l+1)}$$
		$$nDCG@k = \frac{DCG@k}{\sum_{l=1}^{\min(k, ||\mathbf{y}||_0)} \frac{1}{\log(l +1) }}$$
		$$PSnDCG@k = \frac{PSDCG@k}{\sum_{l=1}^{k} \frac{1}{\log l +1 }}$$
Here, $p_l$ is propensity score of the label $l$ proposed in~\citep{Jain16}.

\subsection{Theorem proofs}
\begin{proof} \label{sup:proof}
\textbf{(Bound on $||\hat{\vx}_i - \vv_i||_2$)}. For notational convenience, we use $\hat{\vx} := \hat{\vx}_i$ and $\vv := \vv_i$;
\begin{align*}
    \hat{\vx}_i - \vv_i &= ReLU(\vR \vv) \\
    ||\hat{\vx}_i - \vv_i||_2 &=  ||ReLU(\vR \vv)||_2 
    \shortintertext{\center Using, $||ReLU(\vu)||_2 \le ||\vu||_2$}
    &\le ||\vR \vv||_2
    \shortintertext{\center Using, $||\vR||_{op} \le \lambda$}
    &\le \lambda||\vv||_2 
\end{align*}

\textbf{(Bound on $\hat{\vx}_i$)}.
\begin{align*}
    ||\hat{\vx}_i||_2 &\le ||\vv||_2 + ||\hat{\vx}_i - \vv_i||_2
    \shortintertext{\center Using, $||\hat{\vx}_i - \vv_i||_2 \le \lambda||\vv||_2$}
    &= ||\vv||_2 + \lambda||\vv||_2\\
    &= (1+\lambda)||\vv||_2
\end{align*}

In order to prove bound on cosine similarity, we first prove bound on $||\vmu_l - \vmu^0_l||_2$ and $||\vmu_l||_2$. For notational convenience, we use $\cP := \cP_l$, $\vmu^0 := \vmu^0_l$, and $\vmu := \vmu_l$, \\
\textbf{(Bound on $\vmu_l - \vmu^0_l$)}.
\begin{align*}
    \vmu_l - \vmu^0_l &= \frac{1}{|\cP|} \sum_{i \in \cP}{\hat{\vx}_i - \vv_i} \\
    ||\vmu_l - \vmu^0_l||_2 &= ||\frac{1}{|\cP|} \sum_{i \in \cP}{\hat{\vx}_i - \vv_i}||_2 \\
    &= ||\frac{1}{|\cP|} \sum_{i \in \cP}{ReLU(\vR \vv_i})||_2 \\
    &\le \frac{1}{|\cP|}\sum_{i \in \cP}|| ReLU(\vR \vv_i)||_2
    \shortintertext{\center Using $||ReLU(\vu)||_2 \le ||\vu||_2$}
    &\le \frac{1}{|\cP|}\sum_{i \in \cP}|| \vR \vv_i||_2
    \shortintertext{\center Using, $||\vR||_{op} \le \lambda$}
    &\le  \frac{\lambda}{|\cP|}\sum_{i \in \cP}{|| \vv_i}||_2\\
    \shortintertext{\center Let, $\mathrm{V} := [||\vv_1||_2,..,||\vv_{|\cP|}||_2]$,}
    &= \frac{\lambda}{|\cP|}||\mathrm{V}||_1 
    \shortintertext{\center As, $||\mathrm{V}||_1 \le \sqrt{|\cP|}||\mathrm{V}||_2$}
    &\le \frac{\lambda}{\sqrt{|\cP|}}||\mathrm{V}||_2 \\
    &\le \frac{\lambda}{\sqrt{|\cP|}}[\sum_{i \in \cP}||\vv_i||_2^2]^{\frac{1}{2}} \\
    &= \frac{\lambda}{\sqrt{|\cP|}}[\sum_{i \in \cP}{\vv_i}^T \vv_i]^{\frac{1}{2}}
    \shortintertext{\center As, $\sum_{i \in \cP}\vv_i^T \vv_i \le [\sum_{i \in \cP}\vv_i]^T[\sum_{i \in \cP} \vv_i]$}
    &\le \frac{\lambda}{\sqrt{|\cP|}}\{[\sum_{i \in \cP}\vz_i]^T[\sum_{i \in \cP}\vz_i]\}^{\frac{1}{2}}\\
    &\le \frac{\lambda|\cP|}{\sqrt{|\cP|}}[\vmu^{0{\top}} \vmu^0]^{\frac{1}{2}} \\
    &= \lambda\sqrt{|\cP|}||\vmu^0||_2 
\end{align*}

\textbf{(Bound on $\vmu_l$)}.
\begin{align*}
\vmu &= \frac{1}{|\cP|} \sum_{i \in \cP}{\hat{\vx}_i} \\
||\vmu||_2 &= ||\frac{1}{|\cP|} \sum_{i \in \vP} \hat{\vx}_i||_2 \\
&= ||\frac{1}{|\cP|} \sum_{i \in \cP}{\vv_i + \frac{1}{|\cP|} \sum_{i \in \cP}{\hat{\vx}_i - \vv_i}||_2} \\
&\le ||\frac{1}{|\cP|} \sum_{i \in \cP}{\vv_i}||_2 + ||\frac{1}{|\cP|} \sum_{i \in \cP}{\hat{\vx}_i - \vv_i}||_2 \\
&= ||\vmu^0||_2 + ||\vmu_l - \vmu^0_l||_2 \\
&\le (1+\lambda\sqrt{|\cP|}) ||\vmu^0||_2
\end{align*}

\textbf{(Lower bound on $C(\hat{\vx}, \vmu_l)$)}. 
\begin{align*}
     \frac{C(\hat{\vx}, \vmu)}{C(\vv, \vmu^0)} &= {\frac{\hat{\vx}^{\top}\vmu \cdot ||\vv||_2 ||\vmu^0||_2}{\vv^{\top}\vmu^0 \cdot ||\hat{\vx}||_2 ||\vmu||_2}}
     \shortintertext{\center Using, $||\hat{\vx}||_2 \le (1+\lambda)||\vv||_2$ and, $||\vmu||_2 \le (1+\lambda \sqrt{|\cP|})|| \vmu^0||_2$}
     &\ge \frac{\hat{\vx}^{\top}\vmu}{(1+\lambda)(1+\lambda \sqrt{|\cP|}) \cdot \vv^{\top} \vmu^0}\\
     &\ge \frac{(\hat{\vx})^{\top}(\vmu)}{(1+\lambda \sqrt{|\cP|})^2 \cdot \vv^{\top}\vmu^0}
     \shortintertext{\center Using, $\hat{\vx}^{\top}\vmu \ge \vv^{\top} \vmu^0$, as, $\hat{\vx} - \vv \ge 0$, and $\vmu - \vmu^0 \ge 0$} 
     &\ge \frac{1}{(1+\lambda \sqrt{|\cP|})^2}
\end{align*}

\textbf{(Upper bound on $C(\hat{\vx}_i, \vmu_l)$)}
\begin{align*}
     C(\hat{\vx}, \vmu_l) &= 
     \frac{\hat{\vx}^{\top} \vmu }{||\hat{\vx}||_2 ||\vmu||_2}
     \shortintertext{\center Using, $||\hat{\vx}||_2 \ge ||\vv||_2$ and, $||\vmu||_2 \ge ||\vmu^0||_2$}
     &\le \frac{\hat{\vx}^{\top} \vmu}{||\vv||_2 ||\vmu^0||_2} \\
     &= \frac{\vv^{\top} \vmu^0 + \vv^{\top} (\vmu - \vmu^0) + (\hat{\vx} - \vv)^{\top}\vmu^0 + (\hat{\vx} - \vv)^{\top} (\vmu - \vmu^0)}{||\vv||_2 ||\vmu^0||_2} \\
     &= C(\vv, \vmu^0) \\ 
     &+ \frac{  ||\vv^{\top} \vmu||_2 + ||\vv^{\top}\vmu^0||_2 + ||\hat{\vx}^{\top} \vmu||_2 - ||\vv^{\top} \vmu||}{||\vv||_2 ||\vmu^0||_2} \\
     &\le C(\vv, \vmu^0) \\ 
     &+ \frac{ ||\vv||_2 ||\vmu||_2 + ||\vv||_2 ||\vmu^0||_2 + ||\hat{\vx}||_2 ||\vmu||_2 - ||\vv||_2 ||\vmu||}{||\vv||_2 ||\vmu^0||_2} \\
      \shortintertext{\center Using, $||\hat{\vx}||_2 \le (1+\lambda)||\vv||_2$ and $||\vmu||_2 \le (1+\lambda \sqrt{|\cP|})||\vmu^0||_2$}
      &\le C(\vv, \vmu^0) \\
      &+ \frac{((1+\lambda)(1+\lambda\sqrt{|\cP|})-1)||\vv||_2||\vmu^0||_2}{||\vv||_2 ||\vmu^0||_2} \\
      &\le C(\vv, \vmu^0) + (1+\lambda \sqrt{|\cP|})^2 -1  \qedhere
\end{align*}
\end{proof}

\captionsetup[figure]{position=bottom,justification=centering,width=.85\textwidth,labelfont=bf,font=small}
\begin{figure*}
        \centering
      \begin{subfigure}[t]{0.4\textwidth}
                \includegraphics[width=0.9\linewidth]{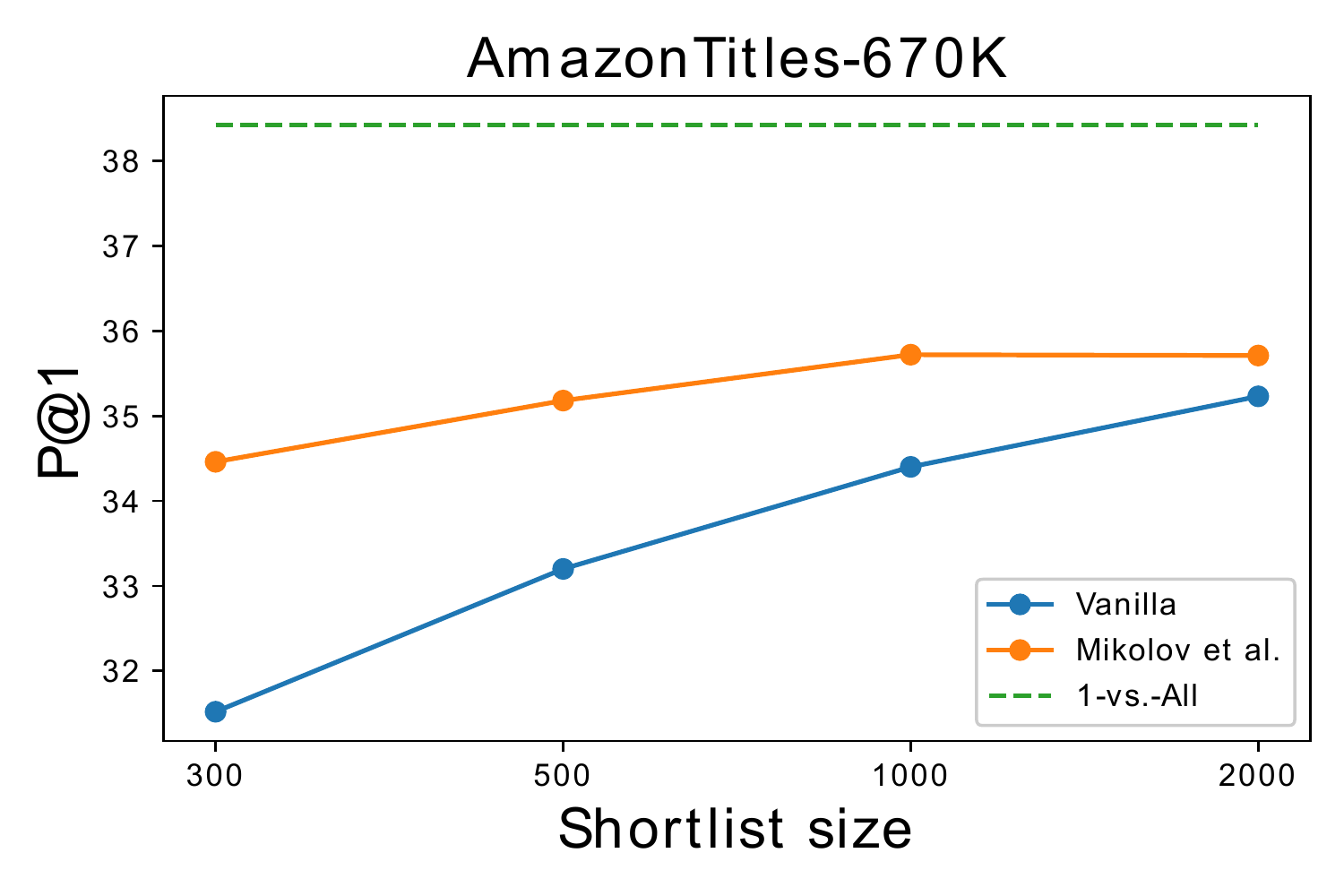}
        \end{subfigure}%
        ~~
        \begin{subfigure}[t]{0.4\textwidth}
                \includegraphics[width=0.95\linewidth]{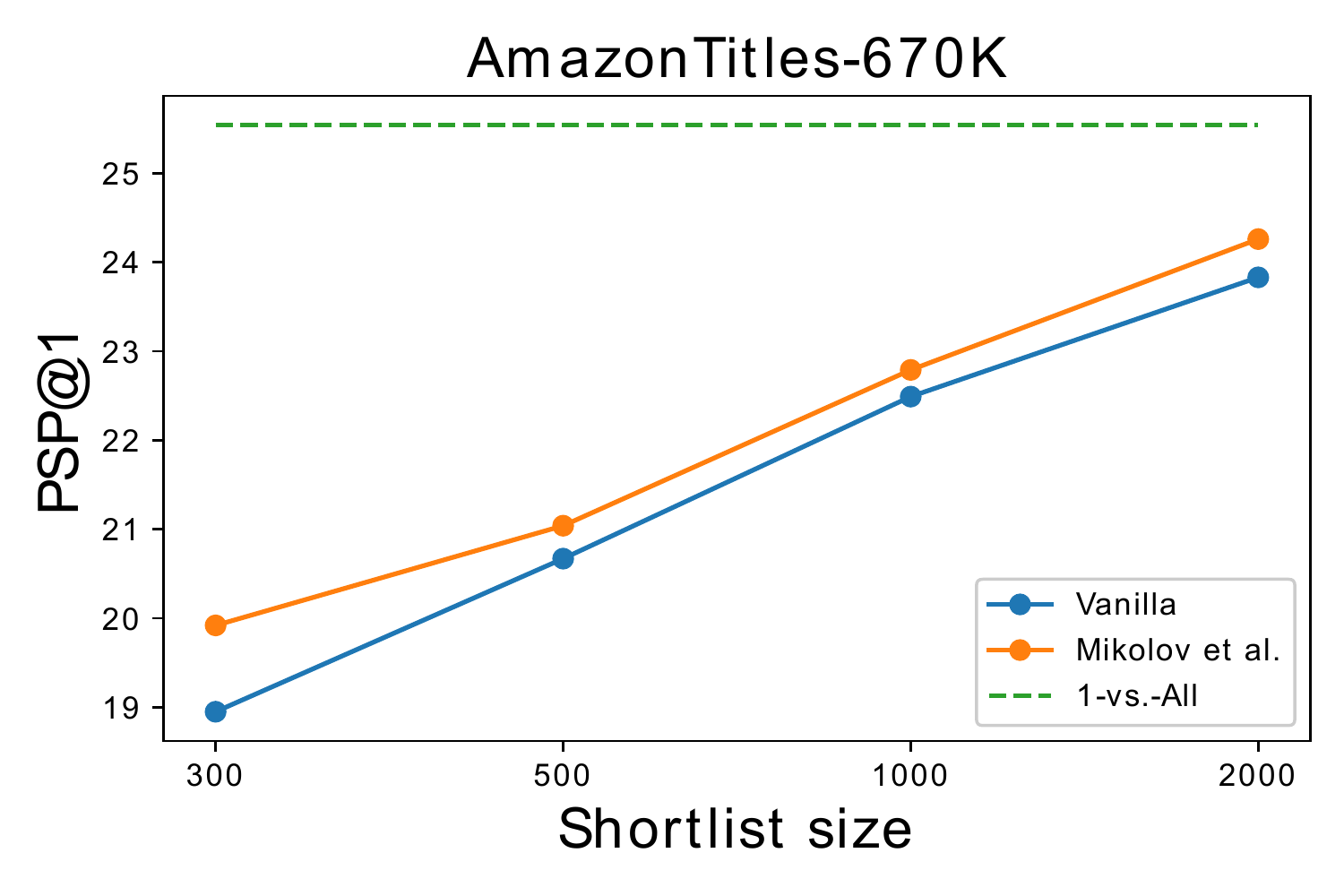}
        \end{subfigure}
        \caption{Performance of DeepXML when trained with a shortlist of randomly sampled negatives as compared to 1-vs.-All strategy. Vanilla strategy samples labels uniformly at random whereas Mikolov et al. samples label based on a unigram distribution over label frequencies. Astec's architecture was used for these experiments}
        \label{fig:supp:negative_sampling}
\end{figure*}

\begin{figure*}
    \centering
    \begin{minipage}{0.4\textwidth}
        \includegraphics[width=0.95\linewidth]{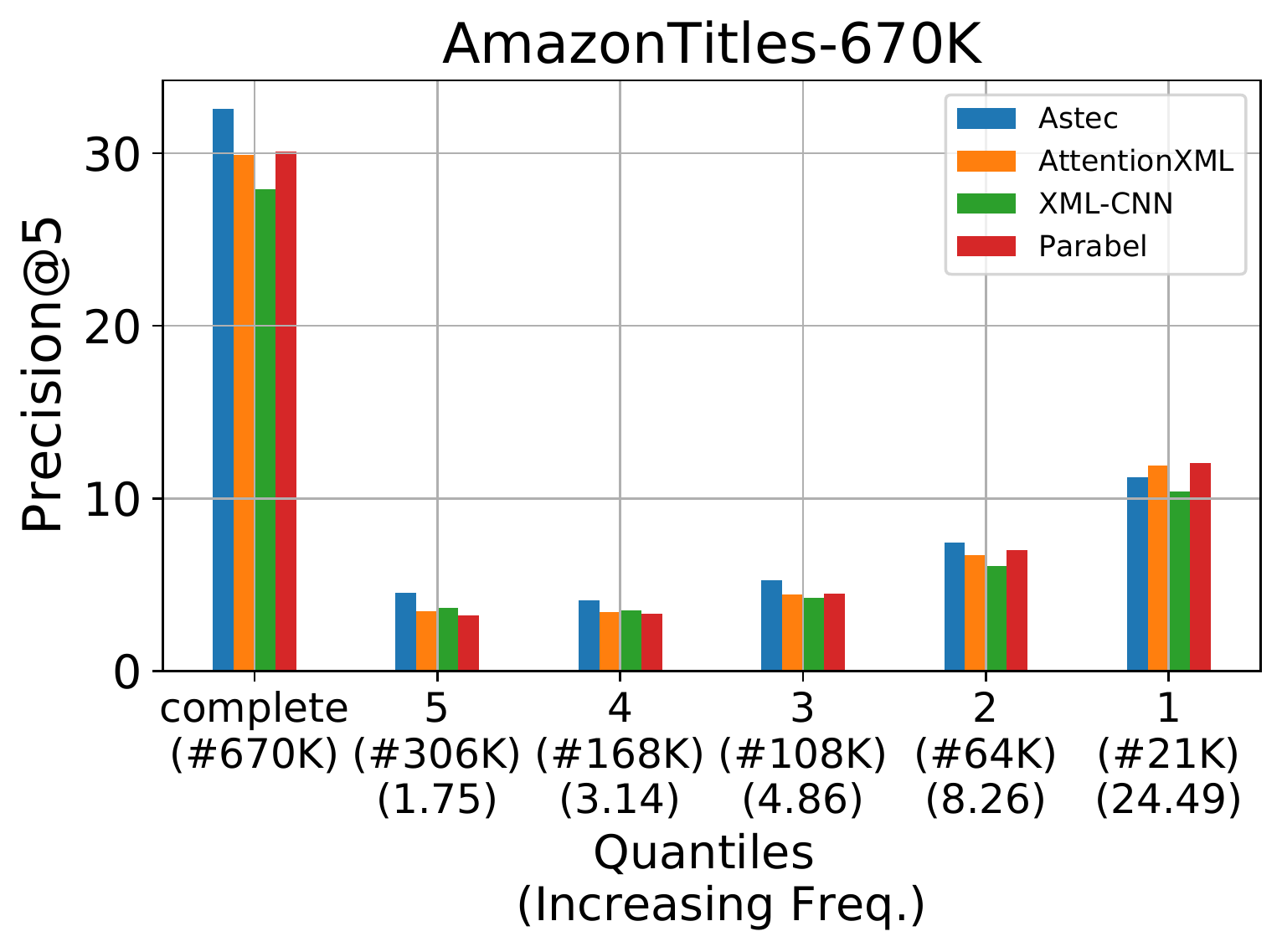}
        \label{fig:AmazonTitles670Ktopkfreqlabels}    
    \end{minipage}~
    \begin{minipage}{0.4\textwidth}
        \includegraphics[width=0.95\linewidth]{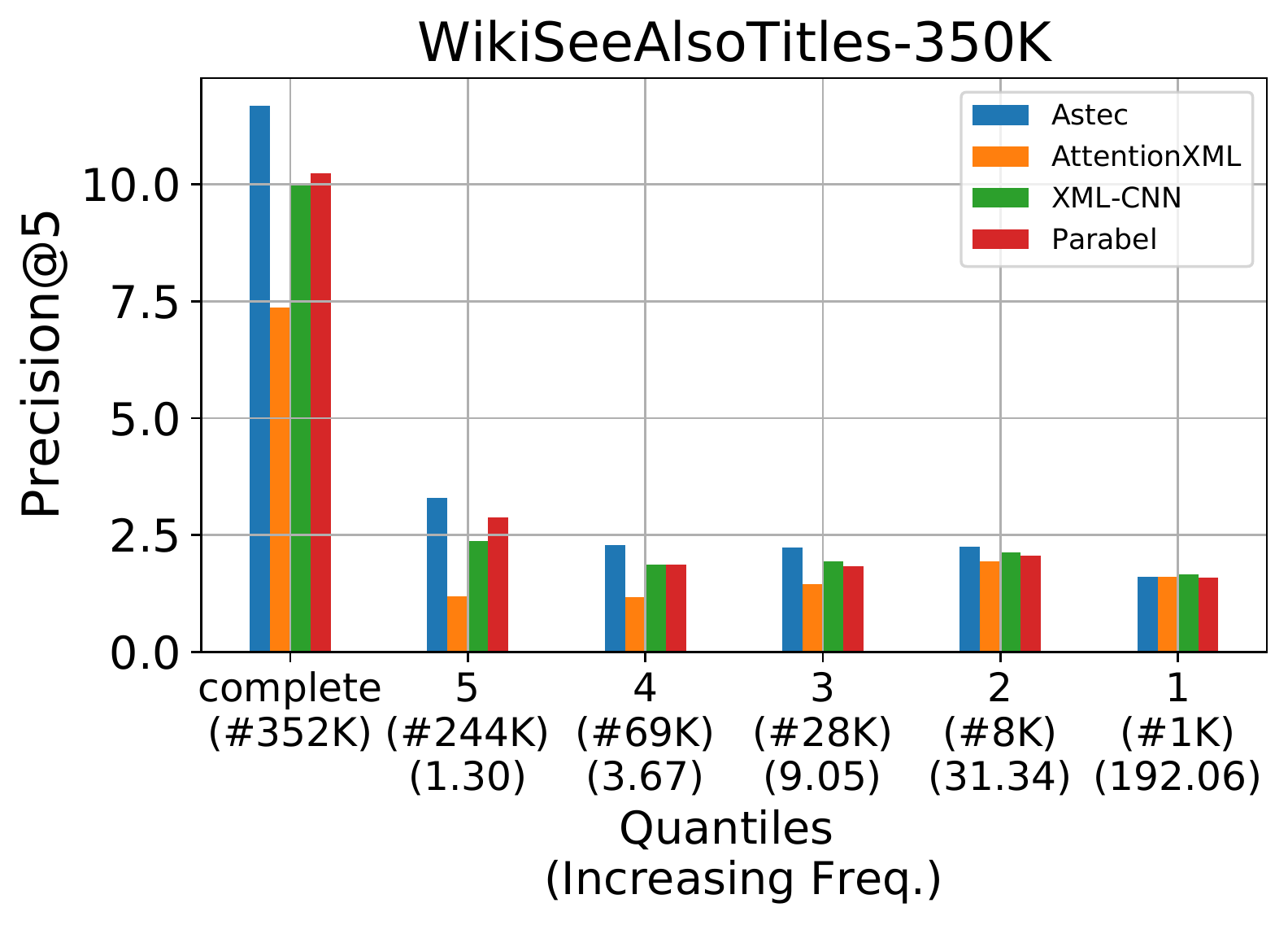}
        \label{fig:contribAmazonTitles670K}
    \end{minipage}
        \caption{Quantile analysis of gains offered by \alg in terms of contribution to P@5 on the WikiSeeAlsoTitles-350K and AmazonTitles-670K datasets. The label set was divided into five equal sized bins~(mean frequency in parenthesis). \alg gains are more prominent on data-scarce tail labels}
\end{figure*}
\begin{table*}
    \small
	\caption{\alg's predicted tags for the Wikipedia title "Confederate Secret Service" are more accurate and diverse as compared to leading methods. All mispredictions have been \textit{italicized}.}
	\label{tab:tab:sup:examples}
      \centering
        \begin{tabular}{@{}ll}
        \toprule
        \textbf{Method} & \textbf{Predictions} \\
        \midrule
        \alg & 1865 disestablishments in the Confederate States of America, Government of the Confederate States of America \\ 
        & 1861 establishments in the Confederate States of America, \textit{Economic history of the Confederate States of America}, \\
        & Military history of the Confederate States of America \\
        \midrule
        \alg (without re-ranker) &  1865 disestablishments in the Confederate States of America, Government of the Confederate States of America \\
        & 1861 establishments in the Confederate States of America, \textit{Economic history of the Confederate States of America} \\
        & \textit{Confederate States of America monuments and memorials}\\
        \midrule
        XML-CNN &  1865 disestablishments in the Confederate States of America, 1861 establishments in the Confederate States of America \\
        & \textit{American films}, \textit{English-language films}, \textit{Black-and-white films}\\
        \midrule
        AttentionXML & \textit{American films}, \textit{English-language films}, Military history of the Confederate States of America, \textit{2011 television episodes} \\ & \textit{English-language television programming} \\
        \bottomrule
        \end{tabular}
\end{table*}

\begin{table*}
    \small
	\caption{\alg's predicted ads for the user Webpage title \& URL Masking Tapes products - ``Grainger Industrial Supply \& https://www.grainger.com/search/adhesives-sealants-and-tape/tapes/masking-tapes" are more accurate and diverse as compared to leading methods~(M1--M2) in Bing.}
	\label{tab:qual_analysis_pads}
      \centering
        \begin{tabular}{@{}ll}
        \toprule
        \textbf{Method} & \textbf{Predictions} \\
        \midrule
        \alg & cheap masking tape, automotive paint masking tape, masking tape in bulk, blue masking tape\\
        & black masking tape, 3 inch masking tape \\
        \midrule
        M1 & 3m reflective tape, safety tape, 3m tape products, 3m packing tape, 3m 250 tape\\
        \midrule
        M2 & online industrial supply, industrial supply company, industrial supply inc, industrial supply \\ 
        & national industrial supply company\\
        \bottomrule
        \end{tabular}
\end{table*}
\begin{table*}
	\caption{Results on AmazonImagesCat-13K dataset}
	\label{tab:resultsvision}
      \centering
        \begin{tabular}{@{}l|ccccc}
        \toprule
        \textbf{Method} & \textbf{P@1} & \textbf{P@3} & \textbf{P@5} & \textbf{N@3} & \textbf{N@5}\\
        \midrule
        DeepXML  &  \textbf{77.19}  &	\textbf{54.8}  &	41.45 &	\textbf{63.48} & \textbf{59.25} \\
        MACH	 &  73.57 &	53.88 &	\textbf{41.99} &	61.8  &	58.76    \\
        Slice    &	48.31 &	35.79 &	27.75 &	41.22 &	39.71 \\
        DiSMEC   &	65.69 &	46.63 &	36.76 &	53.82 &	51.62 \\
        Parabel	 &  64.13 &	45.7  &	36.00 & 52.88 & 50.7 \\
        \bottomrule
        \end{tabular}
\end{table*}

\begin{table}
    \caption{Results on full-text datasets. Please note that `*' marked algorithms uses slightly different version of the dataset. Values indicated by `-' were not available.}
    \label{tab:supp:results_full_text}
      \centering
      \resizebox{\linewidth}{!}{
        \begin{tabular}{@{}l|cccccc|cccccc}
        \toprule
        \textbf{Method} & \textbf{P@1} & \textbf{P@3} & \textbf{P@5} & \textbf{N@1} & \textbf{N@3} & \textbf{N@5} & \textbf{PSP@1} & \textbf{PSP@3} & \textbf{PSP@5} & \textbf{PSN@1} & \textbf{PSN@3} & \textbf{PSN@5}\\
        
        \midrule
        \multicolumn{12}{c}{Wikipedia-500K}\\ \midrule
        \alg & 71.68 & 50.73 & 39.39 & 71.67 & 62.63 & 60.79 & 29.93 & 35.59 & 39.92 & 29.93 & 35.45 & 38.85  \\ 
        \alg-3 & 73.02 & 52.02 & 40.53 & 73.02 & 64.10 & 62.32 & 30.69 & 36.48 & 40.38 & 30.69 & 36.33 & 39.84  \\ 
        XML-CNN & 59.85 & 39.28 & 29.81 & 59.85 & 48.67 & 46.12 & - & - & - & - & - & -  \\ 
        XT & 64.48 & 45.84 & 35.46 & - & - & - & - & - & - & - & - & -  \\ 
        X-Transformer* & 76.95 & 58.42 & 46.14 & - & - & - & - & - & - & - & - & -  \\ 
        AttentionXML & 82.73 & 63.75 & 50.41 & 82.73 & 76.56 & 74.86 & 34.00 & 44.32 & 50.15 & 34.00 & 42.99 & 47.69  \\ 
        SLICE+FastText & 27.98 & 16.61 & 12.11 & 27.98 & 22.81 & 22.69 & 15.04 & 14.61 & 15.17 & 15.04 & 15.97 & 17.59  \\ 
        DiSMEC & 70.20 & 50.60 & 39.70 & 70.20 & 42.10 & 40.50 & 31.20 & 33.40 & 37.00 & 31.20 & 33.70 & 37.10  \\ 
        Parabel & 68.70 & 49.57 & 38.64 & 68.70 & 60.51 & 58.62 & 26.88 & 31.96 & 35.26 & 26.88 & 31.73 & 34.61  \\ 
        AnnexML & 64.64 & 43.20 & 32.77 & 64.64 & 54.54 & 52.42 & 26.88 & 30.24 & 32.79 & 26.88 & 30.71 & 33.33  \\ 
        PfastreXML & 59.50 & 40.20 & 30.70 & 59.50 & 30.10 & 28.70 & 29.20 & 27.60 & 27.70 & 29.20 & 28.70 & 28.30  \\ 
        ProXML & 68.80 & 48.90 & 37.90 & 68.80 & 39.10 & 38.00 & 33.10 & 35.00 & 39.40 & 33.10 & 35.20 & 39.00  \\ 
        Bonsai & 69.20 & 49.80 & 38.80 & 69.20 & 60.99 & 59.16 & 27.46 & 32.25 & 35.48 & - & - & -  \\ 
        \midrule

        \multicolumn{12}{c}{Amazon-670K}\\ \midrule
        Astec & 46.37 & 41.54 & 38.03 & 46.37 & 43.97 & 42.53 & 31.30 & 34.23 & 36.92 & 31.30 & 32.95 & 34.18  \\ 
        Astec-3 & 47.77 & 42.79 & 39.10 & 47.77 & 45.28 & 43.74 & 32.13 & 35.14 & 37.82 & 32.13 & 33.80 & 35.01  \\ 
        XML-CNN & 35.39 & 31.93 & 29.32 & 35.39 & 33.74 & 32.64 & 28.67 & 33.27 & 36.51 &  &  &   \\ 
        XT & 42.50 & 37.87 & 34.41 & 42.50 & 40.01 & 38.43 & 24.82 & 28.20 & 31.24 & 24.82 & 26.82 & 28.29  \\ 
        AttentionXML & 47.58 & 42.61 & 38.92 & 47.58 & 45.07 & 43.50 & 30.29 & 33.85 & 37.13 &  &  &   \\ 
        SLICE+FastText & 33.15 & 29.76 & 26.93 & 33.15 & 31.51 & 30.27 & 20.20 & 22.69 & 24.70 & 20.20 & 21.71 & 22.72  \\ 
        DiSMEC & 44.70 & 39.70 & 36.10 & 44.70 & 42.10 & 40.50 & 27.80 & 30.60 & 34.20 & 27.80 & 28.80 & 30.70  \\ 
        Parabel & 44.89 & 39.80 & 36.00 & 44.89 & 42.14 & 40.36 & 25.43 & 29.43 & 32.85 & 25.43 & 28.38 & 30.71  \\ 
        AnnexML & 42.39 & 36.89 & 32.98 & 42.39 & 39.07 & 37.04 & 21.56 & 24.78 & 27.66 & 21.56 & 23.38 & 24.76  \\ 
        PfastreXML & 39.46 & 35.81 & 33.05 & 39.46 & 37.78 & 36.69 & 29.30 & 30.80 & 32.43 & 29.30 & 30.40 & 31.49  \\ 
        ProXML & 43.50 & 38.70 & 35.30 & 43.50 & 41.10 & 39.70 & 30.80 & 32.80 & 35.10 & 30.80 & 31.70 & 32.60  \\ 
        \midrule
        \bottomrule
    \end{tabular}}
\end{table}

\begin{table}
    \caption{\alg could be significantly more accurate and scalable than leading deep extreme classifiers including MACH, XML-CNN and AttentionXML on publicly available short-text benchmark datasets.}
    \label{tab:results_titles}
      \centering
      \resizebox{\linewidth}{!}{
        \begin{tabular}{@{}l|cccccc|cccccc}
        \toprule
        \textbf{Method} & \textbf{P@1} & \textbf{P@3} & \textbf{P@5} & \textbf{N@1} & \textbf{N@3} & \textbf{N@5} & \textbf{PSP@1} & \textbf{PSP@3} & \textbf{PSP@5} & \textbf{PSN@1} & \textbf{PSN@3} & \textbf{PSN@5}\\

        \midrule
        \multicolumn{12}{c}{WikiSeeAlsoTitles-320K}\\ \midrule
        \alg & 20.42 & 14.44 & 11.39 & 20.42 & 19.90 & 20.63 & 9.83 & 12.05 & 13.94 & 9.83 & 11.67 & 12.90  \\ 
        \alg-3 & 20.61 & 14.58 & 11.49 & 20.61 & 20.08 & 20.80 & 9.91 & 12.16 & 14.04 & 9.91 & 11.76 & 12.98  \\ 
        MACH & 14.79 & 9.57 & 7.13 & 14.79 & 13.83 & 14.05 & 6.45 & 7.02 & 7.54 & 6.45 & 7.20 & 7.73  \\ 
        XML-CNN & 17.75 & 12.34 & 9.73 & 17.75 & 16.93 & 17.48 & 8.24 & 9.72 & 11.15 & 8.24 & 9.40 & 10.31  \\ 
        XT & 16.55 & 11.37 & 8.93 & 16.55 & 15.88 & 16.47 & 7.38 & 8.75 & 10.05 & 7.38 & 8.57 & 9.46  \\ 
        SLICE+fastText & 18.13 & 12.87 & 10.29 & 18.13 & 17.71 & 18.52 & 8.63 & 10.78 & 12.74 & 8.63 & 10.37 & 11.63  \\ 
        AttentionXML & 15.86 & 10.43 & 8.01 & 15.86 & 14.59 & 14.86 & 6.39 & 7.20 & 8.15 & 6.39 & 7.05 & 7.64  \\ 
        DiSMEC & 16.61 & 11.57 & 9.14 & 16.61 & 16.09 & 16.72 & 7.48 & 9.19 & 10.74 & 7.48 & 8.95 & 9.99  \\ 
        Parabel & 17.24 & 11.61 & 8.92 & 17.24 & 16.31 & 16.67 & 7.56 & 8.83 & 9.96 & 7.56 & 8.68 & 9.45  \\ 
        AnnexML & 14.96 & 10.20 & 8.11 & 14.96 & 14.20 & 14.76 & 5.63 & 7.04 & 8.59 & 5.63 & 6.79 & 7.76  \\ 
        PfastreXML & 15.09 & 10.49 & 8.24 & 15.09 & 14.98 & 15.59 & 9.03 & 9.69 & 10.64 & 9.03 & 9.82 & 10.52  \\ 
        Bonsai & 17.95 & 12.27 & 9.56 & 17.95 & 17.13 & 17.66 & 8.16 & 9.68 & 11.07 & 8.16 & 9.49 & 10.43  \\ 
        \midrule

        \multicolumn{12}{c}{AmazonTitles-670K}\\ \midrule
        \alg & 39.97 & 35.73 & 32.59 & 39.97 & 37.91 & 36.60 & 27.59 & 29.79 & 31.71 & 27.59 & 28.80 & 29.61  \\ 
        \alg-3 & 40.63 & 36.22 & 33.00 & 40.63 & 38.45 & 37.09 & 28.07 & 30.17 & 32.07 & 28.07 & 29.20 & 29.98  \\ 
        MACH & 34.92 & 31.18 & 28.56 & 34.92 & 33.07 & 31.97 & 20.56 & 23.14 & 25.79 & 20.56 & 22.18 & 23.53  \\ 
        XML-CNN & 35.02 & 31.37 & 28.45 & 35.02 & 33.24 & 31.94 & 21.99 & 24.93 & 26.84 & 21.99 & 23.83 & 24.67  \\ 
        XT & 36.57 & 32.73 & 29.79 & 36.57 & 34.64 & 33.35 & 22.11 & 24.81 & 27.18 & 22.11 & 23.73 & 24.87  \\ 
        SLICE+fastText & 33.85 & 30.07 & 26.97 & 33.85 & 31.97 & 30.56 & 21.91 & 24.15 & 25.81 & 21.91 & 23.26 & 24.03  \\ 
        AttentionXML & 37.92 & 33.73 & 30.57 & 37.92 & 35.78 & 34.35 & 24.24 & 26.43 & 28.39 & 24.24 & 25.48 & 26.33  \\ 
        DiSMEC & 38.12 & 34.03 & 31.15 & 38.12 & 36.07 & 34.88 & 22.26 & 25.46 & 28.67 & 22.26 & 24.30 & 26.00  \\ 
        Parabel & 38.00 & 33.54 & 30.10 & 38.00 & 35.62 & 33.98 & 23.10 & 25.57 & 27.61 & 23.10 & 24.55 & 25.48  \\ 
        AnnexML & 35.31 & 30.90 & 27.83 & 35.31 & 32.76 & 31.26 & 17.94 & 20.69 & 23.30 & 17.94 & 19.57 & 20.88  \\ 
        PfastreXML & 32.88 & 30.54 & 28.80 & 32.88 & 32.20 & 31.85 & 26.61 & 27.79 & 29.22 & 26.61 & 27.10 & 27.59  \\ 
        Bonsai & 38.46 & 33.91 & 30.53 & 38.46 & 36.05 & 34.48 & 23.62 & 26.19 & 28.41 & 23.62 & 25.16 & 26.21  \\ 
\midrule

        \multicolumn{12}{c}{WikiTitles-500K}\\ \midrule
        \alg & 46.01 & 25.62 & 18.18 & 46.01 & 34.58 & 32.82 & 18.62 & 18.59 & 18.95 & 18.62 & 20.01 & 21.64  \\ 
        \alg-3 & 46.60 & 26.03 & 18.50 & 46.60 & 35.10 & 33.34 & 18.89 & 18.90 & 19.30 & 18.89 & 20.33 & 22.00  \\ 
        MACH & 33.74 & 15.62 & 10.41 & 33.74 & 22.61 & 20.80 & 11.43 & 8.98 & 8.35 & 11.43 & 10.77 & 11.28  \\ 
        XML-CNN & 43.45 & 23.24 & 16.53 & 43.45 & 31.69 & 29.95 & 15.64 & 14.74 & 14.98 & 15.64 & 16.17 & 17.45  \\ 
        XT & 39.44 & 21.57 & 15.31 & 39.44 & 29.17 & 27.65 & 15.23 & 15.00 & 15.25 & 15.23 & 16.23 & 17.59  \\ 
        SLICE+fastText & 28.07 & 16.78 & 12.28 & 28.07 & 22.97 & 22.87 & 15.10 & 14.69 & 15.33 & 15.10 & 16.02 & 17.67  \\ 
        AttentionXML & 42.89 & 22.71 & 15.89 & 42.89 & 30.92 & 28.93 & 15.12 & 14.32 & 14.22 & 15.12 & 15.69 & 16.75  \\ 
        DiSMEC & 39.89 & 21.23 & 14.96 & 39.89 & 28.97 & 27.32 & 15.89 & 15.15 & 15.43 & 15.89 & 16.52 & 17.86  \\ 
        Parabel & 42.50 & 23.04 & 16.21 & 42.50 & 31.24 & 29.45 & 16.55 & 16.12 & 16.16 & 16.55 & 17.49 & 18.77  \\ 
        AnnexML & 39.56 & 20.50 & 14.32 & 39.56 & 28.28 & 26.54 & 15.44 & 13.83 & 13.79 & 15.44 & 15.49 & 16.58  \\ 
        PfastreXML & 30.99 & 18.07 & 13.09 & 30.99 & 24.54 & 23.88 & 17.87 & 15.40 & 15.15 & 17.87 & 17.38 & 18.46  \\ 
        Bonsai & 42.60 & 23.08 & 16.25 & 42.60 & 31.34 & 29.58 & 17.38 & 16.85 & 16.90 & 17.38 & 18.28 & 19.62  \\ 
        \midrule

        \multicolumn{12}{c}{AmazonTitles-3M}\\ \midrule
        \alg & 47.64 & 44.66 & 42.36 & 47.64 & 45.89 & 44.66 & 15.88 & 18.59 & 20.60 & 15.88 & 17.71 & 19.02  \\ 
        \alg-3 & 48.74 & 45.70 & 43.31 & 48.74 & 46.96 & 45.67 & 16.10 & 18.89 & 20.94 & 16.10 & 18.00 & 19.33  \\ 
        MACH & 37.10 & 33.57 & 31.33 & 37.10 & 34.67 & 33.17 & 7.51 & 8.61 & 9.46 & 7.51 & 8.23 & 8.76  \\ 
        XT & 27.99 & 25.24 & 23.57 & 27.99 & 25.98 & 24.78 & 4.45 & 5.06 & 5.57 & 4.45 & 4.78 & 5.03  \\ 
        SLICE+fastText & 35.39 & 33.33 & 31.74 & 35.39 & 34.12 & 33.21 & 11.32 & 13.37 & 14.94 & 11.32 & 12.65 & 13.61  \\ 
        AttentionXML  &  46 & 42.81 & 40.59 & 46.00 & 43.94 & 42.61 & 12.81 & 15.03 & 16.71 & 12.80 & 14.23 & 15.25 \\ 
        Parabel & 46.42 & 43.81 & 41.71 & 46.42 & 44.86 & 43.70 & 12.94 & 15.58 & 17.55 & 12.94 & 14.70 & 15.94  \\ 
        AnnexML & 48.37 & 44.68 & 42.24 & 48.37 & 45.93 & 44.43 & 11.47 & 13.84 & 15.72 & 11.47 & 13.02 & 14.15  \\ 
        PfastreXML & 31.16 & 31.35 & 31.10 & 31.16 & 31.78 & 32.08 & 22.37 & 24.59 & 26.16 & 22.37 & 23.72 & 24.65  \\ 
        Bonsai & 46.89 & 44.38 & 42.30 & 46.89 & 45.46 & 44.35 & 13.78 & 16.66 & 18.75 & 13.78 & 15.75 & 17.10  \\ 
\midrule
        \bottomrule
    \end{tabular}}
\end{table}

\clearpage
\end{document}